\pdfoutput=1

\documentclass[11pt]{article}

\usepackage[preprint]{acl}

\usepackage{times}
\usepackage{latexsym}

\usepackage[T1]{fontenc}

\usepackage[utf8]{inputenc}

\usepackage{microtype}

\usepackage{inconsolata}

\usepackage{graphicx}

\usepackage{hyperref}
\usepackage{rotating}
\usepackage{url}            
\usepackage{booktabs}       
\usepackage{amsfonts}       
\usepackage{nicefrac}       
\usepackage{microtype}      
\usepackage{xcolor} 

\usepackage{colortbl}
\usepackage{pgffor} 
\usepackage{bookmark}

\usepackage{placeins}

\usepackage{enumitem}

\usepackage{lipsum}
\usepackage{graphicx}
\usepackage{amsmath}
\usepackage{etoolbox,xspace}
\usepackage{pdfpages}
\usepackage{breqn}

\usepackage[most]{tcolorbox}
\usepackage{fontawesome}
\usepackage{enumitem}
\usepackage{multirow} 

\usepackage{makecell}
\usepackage{pifont}
\newcommand{\xmark}{\ding{55}}  

\newcommand*{\finmafull}{\texttt{FinMA-7B-full}\xspace}
\newcommand*{\finmanlp}{\texttt{FinMA-7B-NLP}\xspace}
\newcommand*{\gemmanine}{\texttt{Gemma-2-9b}\xspace}
\newcommand*{\gemmanineit}{\texttt{Gemma-2-9b-it}\xspace}
\newcommand*{\llamathreeit}{\texttt{LLaMA-3-8B-Instruct}\xspace}
\newcommand*{\llamathree}{\texttt{LLaMA-3-8B}\xspace}
\newcommand*{\gemmats}{\texttt{Gemma-2-27b}\xspace}
\newcommand*{\gemmatsit}{\texttt{Gemma-2-27b-it}\xspace}
\newcommand*{\llamasit}{\texttt{LLaMA-3-70B-Instruct}\xspace}
\newcommand*{\llamas}{\texttt{LlaMA-3-70B}\xspace}

\newcommand*{\zero}{\textit{zero} model\xspace}

\newcommand*{\one}{\textit{one} model\xspace}

\newcommand*{\random}{\textit{Random} model\xspace}

\newcommand*{\logistic}{\textit{Logistic Regression} model\xspace}

\newcommand*{\great}{\texttt{GReaT}\xspace}

\newcommand*{\html}{\texttt{HTML}\xspace}

\newcommand*{\json}{\texttt{JSON}\xspace}

\newcommand*{\lift}{\texttt{LIFT}\xspace}

\newcommand*{\List}{\texttt{List}\xspace}

\newcommand*{\latex}{\texttt{Latex}\xspace}

\newcommand*{\Text}{\texttt{Text}\xspace}

\newcommand{\myverb}[1]{ \indent{ \begin{verbatim} #1 \end{verbatim} } }
\newtheorem{definition}{Definition}

\title{Accept or Deny? Evaluating LLM Fairness and Performance in Loan Approval across Table-to-Text Serialization Approaches}

\author{
 \textbf{Israel Abebe Azime \thanks{Equal contribution} \textsuperscript{1}},
 \textbf{Deborah D. Kanubala \footnotemark[1] \textsuperscript{1}},
 \textbf{Tejumade Afonja \footnotemark[1] \textsuperscript{1, 2}},
 \textbf{Mario Fritz\textsuperscript{1, 2}},
\\
 \textbf{Isabel Valera\textsuperscript{1,3}},
 \textbf{Dietrich Klakow\textsuperscript{1}},
 \textbf{Philipp Slusallek\textsuperscript{1}}
\\
 \textsuperscript{1} Saarland University,
 \textsuperscript{2} CISPA Helmholtz Center for Information Security, 
 \\
 \textsuperscript{3}Max Planck Institute for Software Systems
}

\begin{document}
\maketitle
\begin{abstract}



Large Language Models (LLMs) are increasingly employed in high-stakes decision-making tasks, such as loan approvals. While their applications expand across domains, LLMs struggle to process tabular data, ensuring fairness and delivering reliable predictions.
In this work, we assess the performance and fairness of LLMs on serialized loan approval datasets from three geographically distinct regions: Ghana, Germany, and the United States. Our evaluation focuses on the model's zero-shot and in-context learning (ICL) capabilities. 
Our results reveal that the choice of \textit{serialization}\footnote{Serialization refers to the process of converting tabular data into text formats suitable for processing by LLMs.} format significantly affects both performance and fairness in LLMs, with certain formats such as \great and \lift yielding higher F1 scores but exacerbating fairness disparities. Notably, while ICL improved model performance by 4.9-59.6\% relative to zero-shot baselines, its effect on fairness varied considerably across datasets. 
Our work underscores the importance of effective tabular data representation methods and fairness-aware models to improve the reliability of LLMs in financial decision-making.

\end{abstract}
\section{Introduction}


Large Language Models (LLMs), trained on vast amounts of textual data, have demonstrated remarkable potential to generalize across tasks and provide accurate predictions~\cite{naveed2023comprehensive, ai4science2023impact}.
Given their growing presence in critical domains like financial decision-making, it is crucial to understand the behaviour and ethical implications of these systems due to their direct and severe impact on individuals~\cite{aguirre-etal-2024-selecting}.
Financial decision-making is the systematic process of analyzing 
information to make informed choices in financial tasks such as investment, loan approval, and more~\cite{kazemian-etal-2022-taxonomical}.

In this work, we focus on loan approval, where a bank must decide whether or not to grant a loan based on the applicant's creditworthiness.
This task is typically performed by loan officers who consider various input factors to make informed decisions. Loan approval is a critical task to explore as it directly impacts financial inclusion, borrower outcomes, and institutional risk management, making it an ideal domain for assessing the effectiveness and fairness of LLM-driven decision-making systems. 
Moreover, given the diversity in financial practices and socioeconomic contexts, evaluating loan approval across datasets from three distinct geographical regions (Ghana, Germany, and the United States) provides valuable insights into how LLMs manage data diversity and fairness within varying economic environments.
Additionally, the tabular nature of the datasets in this study underscores the importance of selecting an appropriate serialization method before feeding data into LLMs, as it can significantly influence model performance and fairness~\cite{singha2023tabular, sui2024table}.  

\begin{figure*}
\centering
\includegraphics[width=0.8\linewidth]{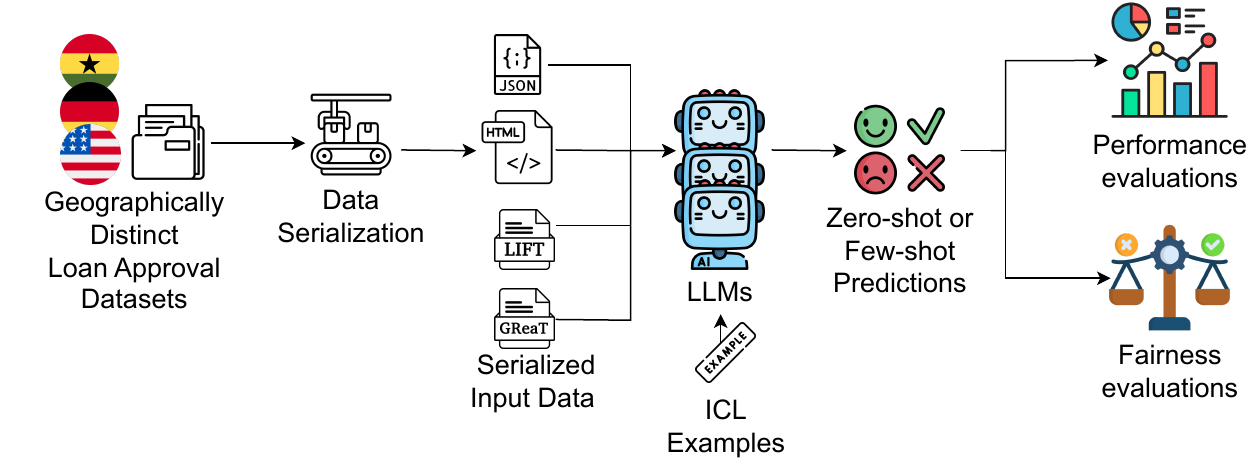}

\caption{\textbf{Overview of our approach.} We first utilize different serialization approaches to acquire our serialized data, and we investigate the LLMs' performance and fairness by applying zero- and few-shot learning to the datasets.
}
    \label{fig:zero-shot-format-serialization}

\end{figure*}

Building upon these observations, we frame our study around the following research questions:
i) How do different serialization formats (e.g., \json, \Text, \great, \lift) impact the fairness and performance of LLMs in loan approval tasks across diverse geographical datasets?
ii) What effect does in-context learning (ICL) have on the fairness and predictive performance of LLMs in loan approval scenarios, particularly when applied to datasets from Ghana, Germany, and the United States?
iii) How do financial domain-specific LLMs compare to general-purpose LLMs in their ability to accurately and fairly assess loan applications, especially under zero-shot and few-shot learning settings?
iv) What key factors contribute to fairness disparities in LLM-generated loan approval predictions, and how do these factors vary across different serialization methods and geographical regions?

\noindent
To address the research questions outlined above, this work makes the following contributions\footnote{\href{https://huggingface.co/collections/israel/accept-or-deny-68ac612eb9bc800b65ef17ce}{Access to the resources on huggingface}

}:

\begin{enumerate}[nosep]
    \item Investigate the capability of LLMs in financial decision-making, focusing on loan approval tasks. This includes a comprehensive zero-shot benchmark evaluation of various LLMs and an analysis of the features they prioritize in their decision-making process.

    \item Analyze the impact of different tabular serialization formats on the decision-making process of LLMs.   
   
    \item Evaluate the effectiveness of techniques, such as in-context learning, that aim to improve LLM performance in financial decision-making, with particular attention to their impact on accuracy and fairness.

    \item Examine the presence of gender-related biases in LLM-generated financial decisions, assessing their implications and associated risks.

\end{enumerate}

\begin{table}[!ht]
\centering
\begin{tabular}{llll}
\toprule
\textbf{Data Name}  & \textbf{Size} & \textbf{\#Features} & \textbf{Output} \\ \midrule
Ghana  & 614 & 13 & Yes/No\\ 
Germany  & 1000 & 21 & Good/Bad \\  
United States & 1451 & 18 & Yes/No \\ 
\bottomrule
\end{tabular}

\caption{\textbf{Summary of the datasets used in the study.} Ghana \cite{sackey2018gender}, Germany \cite{statlog_website} and  United States \cite{kaggle_loan_approval}. See Appendix~\ref{data-description} for details of the feature description of each dataset. }
\label{data_summary}
\end{table}

\section{Related Work}
\label{relatedwork}

\begin{table*}[!ht]
\footnotesize
\centering
\begin{tabular}{l|p{8cm}}
\toprule
\textbf{Serialization}& \textbf{Example Template} \\
\midrule
\texttt{JSON (default)} & 

 \{\texttt{age: 32, sex: female, loan duration: 48 months,
 purpose: education}\} 
  \\
\midrule
\texttt{GReaT \cite{borisov2022language}} & \texttt{age is 32, sex is female, loan duration is 48 months, loan purpose is education} \\
\midrule
\texttt{LIFT \cite{dinh2022lift}}& \texttt{A 32-year-old female is applying for a loan for 48 months for education purposes.}\\
\bottomrule 
\end{tabular}
\caption{\textbf{Comparison of serialization formats for loan applicant information.} This table presents example templates for representing loan applicant data with four features (age and sex, loan duration and purpose). \json is assumed as the default format. 
Table~\ref{example_serialisation_full} in Appendix~\ref{app:serialization} shows examples for the \List, \Text, \html and \latex format. \label{example_serialisation}
}
\end{table*}


\paragraph{LLMs in financial decision-making.} 

Large Language Models (LLMs) have been employed to support various financial decision-making tasks, encompassing diverse applications such as stock trading~\cite{ding2024large}, investment management~\cite{kong2024large}, and credit scoring~\cite{feng2023empowering}. These models either provide recommendations on optimal investment strategies to maximize returns or assess an individual’s financial reliability and creditworthiness~\cite{haque2024bank}. Loan approval tasks, in particular, carry significant risk due to their direct impact on financial inclusion and access to capital, making the evaluation of fairness and predictive accuracy in such models critically important~\cite{kanubala2024fairness}.
\paragraph{Serialization in LLMs.} 
LLMs require tabular data to be serialized into natural text, a process known as serialization~\cite{jaitly2023towards}.
However, serialization methods, which convert tabular data into a format that LLMs can process, can introduce their own biases and limitations. For instance,~\citet{hegselmann2023tabllm} discusses how different serialization formats can lead to variations in LLMs' performance. Their study highlights that the choice of serialization method can influence how effectively an LLM understands and processes the data. A number of studies have proposed different serialization methods, including~\citet{hegselmann2023tabllm} \Text and \List formats, 
the \great format~\cite{borisov2022language},
natural-like serialization as used in \lift~\cite{dinh2022lift}, and \html-like formatting~\cite{sui2024table}. Additionally, works like~\citet{hollmann2022tabpfn} introduce TabPFN, a tabular foundation model specifically designed for tabular datasets. However, in this work, we focus on the capabilities of general-purpose LLMs and their financial domain variants. We do not cover tabular foundation models due to the broad range of serialization formats considered in our study, which may not align well with such models.

\paragraph{Bias and unfairness of LLMs.} LLMs are trained on large corpora of human-generated text, which often contain inherent societal biases~\cite{garg2018word, navigli2023biases, sun2019mitigating, kotek2023gender}. As a result, these biases can be encoded into the models and perpetuated in their decisions, leading to discriminatory outcomes. For instance, gender or racial biases present in the training data can result in unfair treatment of certain groups~\cite{bolukbasi2016man, abid2021persistent}. Additionally,~\citep{aguirre-etal-2024-selecting} highlights that the choice of in-context examples significantly influences model fairness, particularly when these examples are not demographically representative.  Addressing these biases is crucial to ensuring fair and ethical use of LLMs in decision-making processes.

Our study examines the use of LLMs for loan approval decisions across datasets from three geographical regions. We explore two key dimensions: the impact of serialization methods and the effect of zero-shot and few-shot prompting on decision accuracy and fairness. 




\section{Methodology}
\label{methodology}
\subsection{Problem Formalization}
Given the tabular dataset $ D = \{(x_i, y_i)\}_{i=1}^{n} $, where $ x_i $ is a $d$-dimensional feature vector and $y_i$ belongs to a set of classes $C$, the columns or features are named $F = \{f_1, \ldots, f_d\}$. Each feature $f_i$ is a natural-language string representing the name of the feature, such as ``age'' or ``sex''. For zero-shot learning, we provide the LLMs with features $F$ and task it to predict the class $C$. For our k-shot classification experiments, we use a subset $D_k$ of size $k-$sampled from the training set. Few-shot examples are top-n examples balanced by gender to align with fairness metrics.



\subsection{Datasets}

\paragraph{Dataset choice.}
Guided by data availability and relevance, we selected three distinct datasets representing the region’s socioeconomic context. We posit that geographical, political, and ideological differences across regions directly influence financial practices, such as loan acquisition. The regions examined were arbitrarily chosen for this study; while expanding to more diverse regions is feasible, we have limited our scope to maintain a focused analysis. The distinct differences in data properties highlight the geographical variations central to this study. Although the task remains the same, subtle disparities within datasets from specific groups may introduce biases that can impact decision-making.


A comparison of dataset characteristics reveals distinct patterns across the German, Ghanaian, and U.S. datasets, as further detailed in the Appendix~\ref{sec:data-descriptions}. Only the Germany and Ghana datasets include age as a feature, with German applicants predominantly in their 20s and Ghanaian applicants in their 40s. The U.S. dataset primarily emphasizes employment status, whereas the other datasets provide additional information on the number of years employed. Across all datasets, male applicants consistently outnumber female applicants. Notable variations are also observed in loan amount distributions: the Germany dataset presents a broader and more evenly distributed range of loan amounts, while the U.S. and Ghana datasets are concentrated on smaller loan amounts with higher frequency.


\paragraph{Data processing.}  We provide a summary of the dataset we used in the study in Table~\ref{data_summary} with a detailed description in Appendix~\ref{data-description}.
For each dataset, we split the dataset into 80\% train and 20\% test using stratified sampling based on gender feature. 
To convert each dataset to the formats shown in Table~\ref{example_serialisation} we created custom functions and also used pandas \footnote{https://pandas.pydata.org/} functions that change dataframe to \html and \latex. See Table~\ref{example_serialisation_full} in Appendix~\ref{app:serialization} for examples of \latex, \Text, \html and \List formats. 


\subsubsection{Table-to-Text Serialization} 
Converting tabular data to text (\emph{serialization}) is essential, as the format can significantly influence LLM decision-making~\citep{hegselmann2023tabllm}.
To investigate how this behaviour transfers to our loan approval task, we explored \emph{six} serialization formats as shown in Table~\ref{example_serialisation} and Table~\ref{example_serialisation_full} in Appendix~\ref{app:serialization}. These formats ranged from straightforward default values, such as \json and \List, to more structured and natural language text-like formats, such as \html, \latex, \Text~\citep{hegselmann2023tabllm}, \great~\citep{borisov2022language} and \lift~\citep{dinh2022lift}.

\begin{figure*}[!h]
\centering
\includegraphics[width=\textwidth]{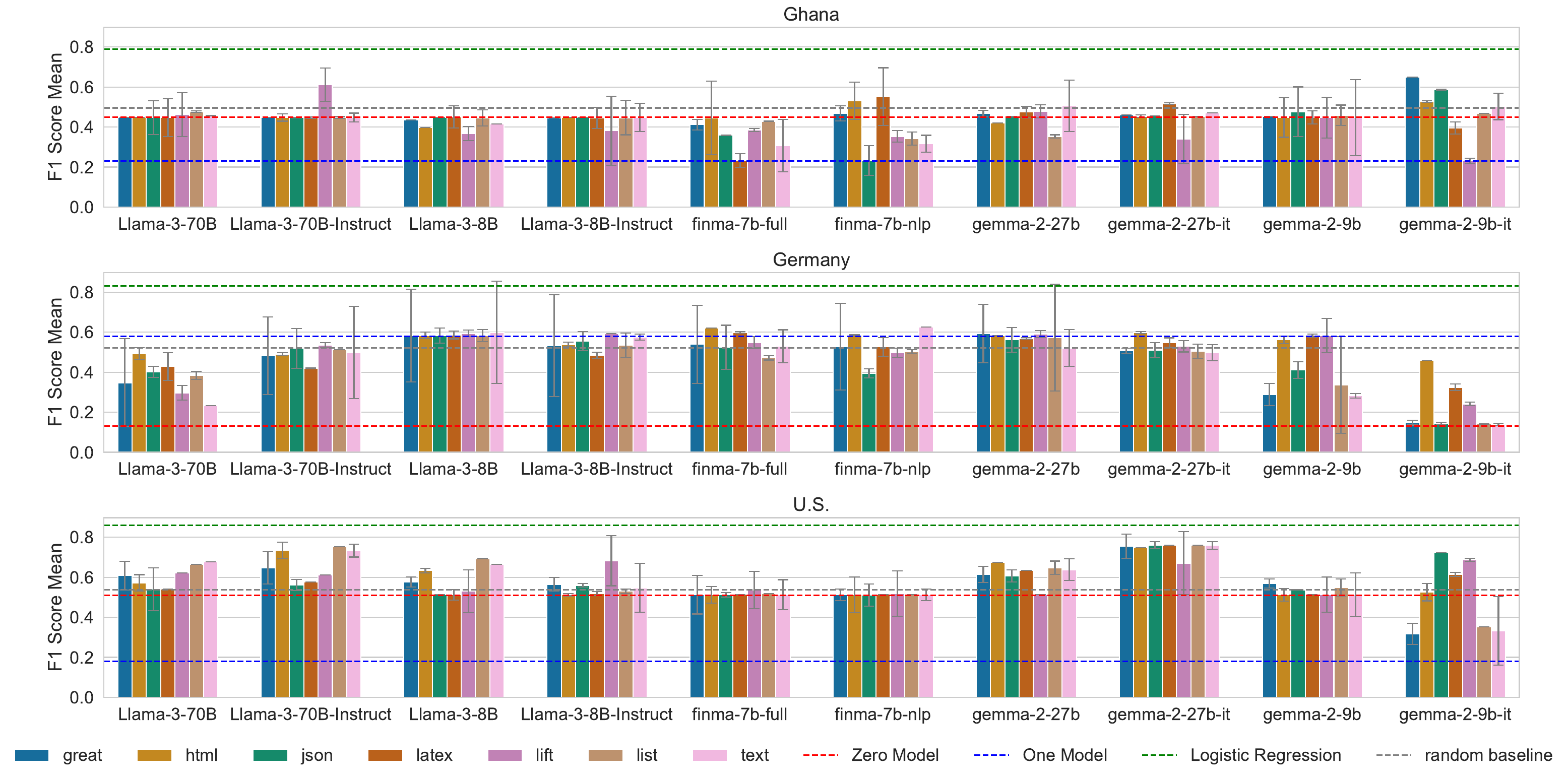}
\caption{\textbf{Zero-shot weighted average F1 score performance of LLMs on loan approval tasks.} Evaluated across three prompts (variation shown by error bars) and multiple table-to-text serialization methods. The \logistic baseline (green dashed line) uses default \json serialization with variables as individual features. Most LLMs underperform relative to this baseline, with only \great on Ghana, \List/\Text on Germany, and \gemmatsit on the U.S. showing modest improvements.}

    \label{fig:zero-shot-format-serialization}
\end{figure*}

\subsection{Models}
\subsubsection{Baseline and Benchmark Models}
To comprehensively understand and accurately evaluate the investigated LLMs, we incorporated simple baseline models and a benchmark model.
\paragraph{Baseline models.}
The \zero, \one and \random serve as our simple baselines, as shown in Figure~\ref{fig:zero-shot-format-serialization}. The \zero assumes that no one will repay the loan (i.e. zero output for all predictions), while the \one assumes that everyone will repay the loan (one output for all predictions). These models provide initial reference points for our experiment, illustrating the performance metrics under these extreme assumptions. Finally, the \random serves as a baseline by comparing the model's performance against randomly generated predictions\footnote{We use NumPy with a fixed seed for reproducibility}.

\paragraph{Benchmark model.}
We trained a \logistic on the training set to serve as our benchmark model. This model allows us to compare the performance of the LLMs against traditional and well-understood machine learning models. In training the \logistic, we preprocessed the dataset by dropping missing values, applying label encoder to the categorical features, and scaling all numerical features using a standard scaler. Additionally, we used default parameters of scikit-learn\footnote{https://scikit-learn.org/} implementation for logistic regression to be used as basic comparison baseline. 
We acknowledge that other classical models, such as decision trees or support vector machines, might be optimized for this task and potentially yield better performance. However, our primary objective was to establish a straightforward benchmark for comparison.

\subsubsection{Large Language Models (LLMs)} 
We evaluated a total of ten (10) LLMs selected based on their open-source availability, instruction tuning, parameter size, and domain relevance (Table~\ref{tab:llm_used}). 
To assess the effect of domain relevance, we included models specifically fine-tuned for financial tasks: \finmanlp and \finmafull, introduced by~\citet{xie2023pixiu}. To examine the effect of instruction tuning, we incorporated Meta’s \llamasit and \llamathreeit, as well as Google’s \gemmatsit and \gemmanineit. Each of these instruction-tuned variants was paired with its corresponding base model (\llamas, \llamathree, \gemmats, and \gemmanine) sourced from~\citet{touvron2023llama, metaIntroducingMeta, gemmateam2024gemma}. This selection allows us to examine both the impact of instruction tuning and the role of model size, while also testing whether financial fine-tuning improves decision-making in domain-specific tasks such as loan approval. 
See Appendix~\ref{eval-setup} for model evaluation setup.



\begin{table}[h]
\centering
\footnotesize
\begin{tabular}{lllc}
\hline
\textbf{Model} & \textbf{Training} & \textbf{Params} & \makecell{\textbf{Financial} \\ \textbf{Dataset} \\ \textbf{Only}} \\
\midrule
\texttt{LLaMA-3} & \begin{tabular}[c]{@{}l@{}}Pretrained \&\\ Instruction-tuned\end{tabular} & 8B \& 70B & \xmark \\
\texttt{Gemma-2} & \begin{tabular}[c]{@{}l@{}}Pretrained \&\\ Instruction-tuned\end{tabular} & 9B \& 27B & \xmark \\
\hline
FinMA-full & Fine-tuned & 7B & \checkmark \\
FinMA-NLP & Fine-tuned & 7B & \checkmark \\
\hline
\end{tabular}
\caption{Overview of the LLMs evaluated, including models fine-tuned and whether they were specifically trained on financial datasets or not.}
\label{tab:llm_used}
\end{table}

\subsection{Approaches to LLMs Improvement}
\subsubsection{In-Context Learning (ICL)} In-context learning involves providing examples that enhance the capabilities of LLMs~\cite{zhang2024impact,agarwal2024many}. This approach is widely used because it eliminates the need for parameter updates, reducing computational costs associated with training. Following a similar approach utilized by the work of~\citet{zhang2024impact}
we experimented with different numbers of examples, specifically \( n = 2, 4, 6, 8 \). Our few-shot examples are strategically selected to ensure representational equity. For instance, when using two examples, one will correspond to a male and the other to a female, aligning with our fairness score metrics, which are based on gender representation.

\subsection{Model and Fairness Evaluation}
We use the weighted-average F1 score to evaluate model performance on the loan prediction task (see Appendix~\ref{metrics} for definitions).
To assess fairness, we employ two standard metrics: equality of opportunity (EO) and statistical parity (SP). EO aligns with the goals of loan approval by ensuring that qualified applicants, regardless of group membership, have an equal chance of approval~\cite{hardt2016equality, kozodoi2022fairness}. In contrast, SP measures whether approval rates are independent of sensitive attributes~\cite{dwork2012fairness}. Formal definitions of these metrics are provided below:

\begin{definition}[Statistical Parity (SP)]
\label{dp-definition} A trained classifier's predictions $\hat{Y}$ satisfy Statistical Parity if the probability of a positive outcome is independent of the sensitive attribute~\cite{dwork2012fairness}. Formally:
\begin{equation*}
P(\hat{Y}=1 \mid A=1) = P(\hat{Y}=1 \mid A=0)
\end{equation*}
where $A$ denotes the sensitive attribute, which we consider to represent \emph{gender}. For simplicity, we assume $A$ is binary: $A \in \{\text{male}, \text{female}\}$. 
Here, $\hat{Y}$ is the \emph{predicted label} of the classifier, and $Y$ denotes the \emph{true target label}.
\end{definition}



\begin{definition}[Equality of Opportunity (EO)]
\label{eo-definition}
Equality of Opportunity ensures that the classifier's \emph{true positive rate} is the same across different demographic groups~\cite{hardt2016equality}. Formally, a classifier $\hat{Y}$ satisfies Equality of Opportunity if:
\begin{multline*}
P(\hat{Y}=1 \mid Y=1, A=1) \\
= P(\hat{Y}=1 \mid Y=1, A=0)
\end{multline*}
where $A$ is the sensitive attribute. For our experiments, we consider females as the protected group and males as the non-protected group.
\end{definition}

\begin{figure*}[!ht]
\centering
\includegraphics[width=0.9\linewidth]{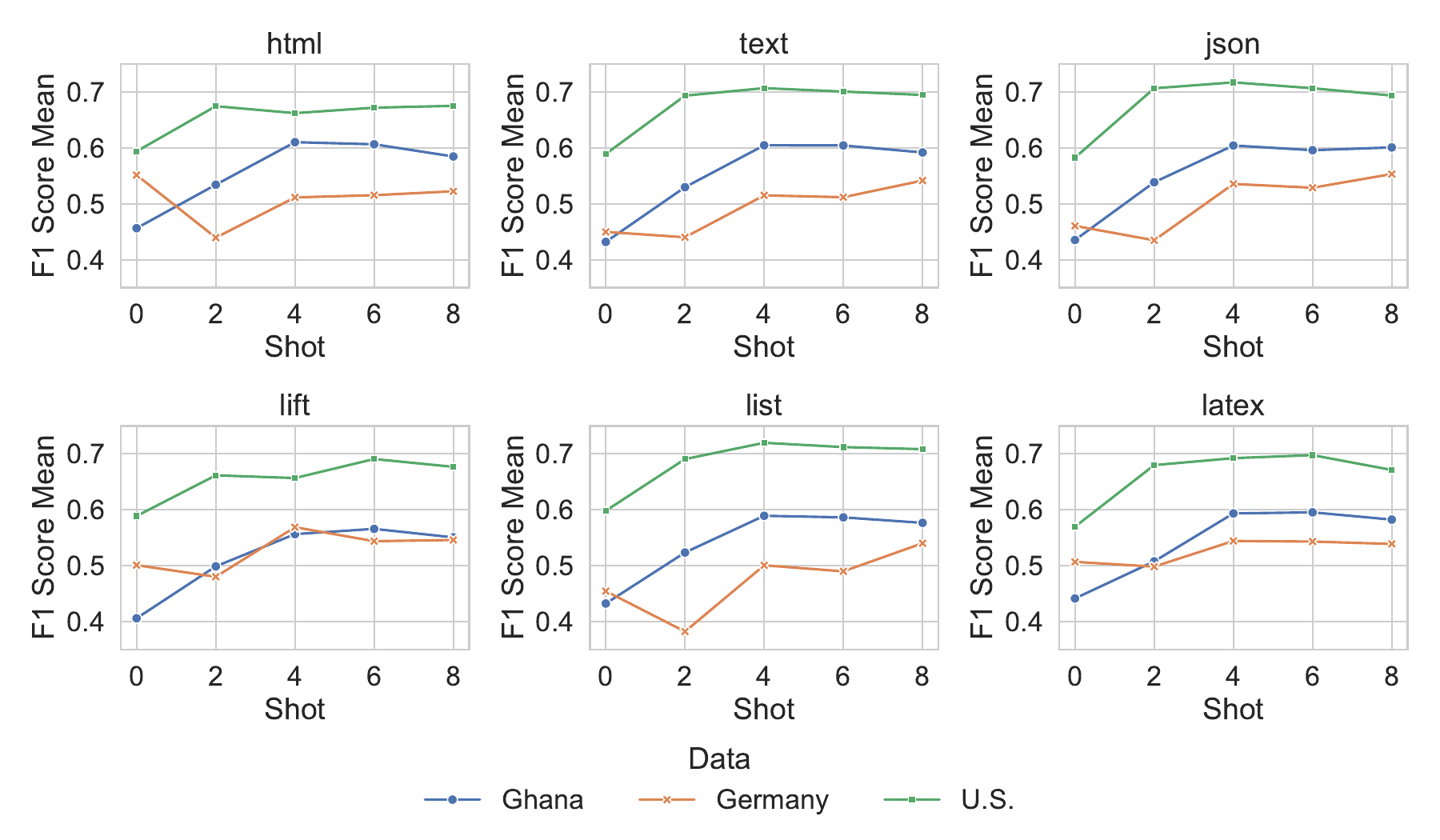}
\caption{
\textbf{Average weighted F1 score} trends across serialization formats for few-shot examples, showing higher gains in U.S. data across formats, while Germany lags consistently despite increasing shot numbers.
}
\label{fig:shot_v_f1_per_data}
\end{figure*}

\section{Results and Analysis}
\label{results}
In this section, we present our results and analysis, structured around a set of research questions that guide the discussion. We begin by comparing the performance of different serialization methods across models for each dataset, as shown in Figure~\ref{fig:zero-shot-format-serialization}. We observe that the \zero outperforms the \one on the Ghana and United States (US) datasets, while the reverse is true for the Germany dataset. This suggests that the Germany dataset has a higher proportion of non-defaulters compared to the other two datasets. We also conducted experiments on model token attribution, which is detailed in the Appendix \ref{token-attrbution-experiments}.
\vspace{-0.25cm}
\subsection{Do LLMs Perform Better Than Baseline or Benchmark Models on the Default Serialization Format (\json)?}
In Figure~\ref{fig:zero-shot-format-serialization}, we compare the zero-shot performance of LLMs against baseline models and analyse the results by country. The general trend indicates that most models do not outperform either the \zero or the \one. Some models achieved marginally higher F1 scores, including \gemmanineit for Ghana and seven of the models for the US, while none did so for Germany. Importantly, none of the selected LLMs were able to outperform the simple \logistic, which serves as the benchmark.



\begin{tcolorbox}[
    colback=blue!3!white,
    colframe=blue!50!black,
    boxrule=0.5mm,       
    arc=2mm,             
    left=1mm,            
    right=1mm,           
    top=1mm,             
    bottom=1mm           
]
    \fontsize{10}{12}\selectfont 
    \faLightbulbO~For \json serialization method, financial domain-specific models (\finmafull, \finmanlp) do not demonstrate significantly better performance under zero-shot decision-making compared to models trained for general applications. Also, none of the models outperform the \logistic. 
\end{tcolorbox}

\subsection{How Does the Zero-Shot Performance of LLMs Vary Across Different Serialization Methods Compared to Baseline Models?}
\label{sec:zero-shot-performance}
Examining region-specific results, we observe the following from Figure~\ref{fig:zero-shot-format-serialization}:
For the \emph{Ghana} dataset, the best performances are achieved using the \great (\gemmanineit) and \lift (\llamasit) serialization method. In the \emph{Germany} dataset, \gemmanineit shows the poorest performance, with three out of four models performing as poorly as the zero model. Financial domain-trained models (\finmafull and \finmanlp) deliver the best results with \List and \Text serialization methods. For the \emph{U.S.} dataset, results are generally more promising across all models, with \gemmatsit consistently achieving the best performance across all serialization methods tested except \lift.

\begin{figure*}[!h]
\centering
\includegraphics[width=\linewidth]{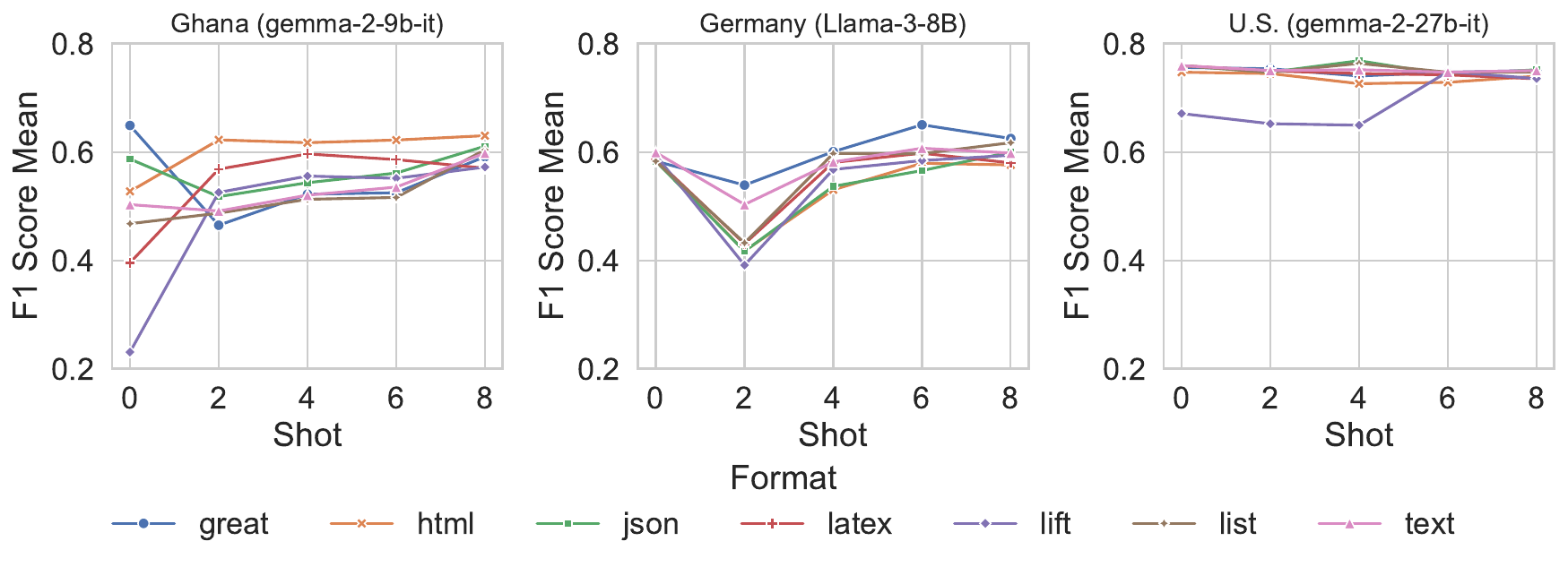}
\caption{
\textbf{Few-shot weighted F1 trends.} Adding a small number of in-context examples improves performance, while differences among serialization formats remain modest across datasets.
}

\label{per_format_score_1x3_shared_legend}
\end{figure*}



\begin{tcolorbox}[
    colback=blue!3!white,
    colframe=blue!50!black,
    boxrule=0.5mm,       
    arc=2mm,             
    left=1mm,            
    right=1mm,           
    top=1mm,             
    bottom=1mm                
]
    {\fontsize{10}{20}\selectfont \faLightbulbO}\hspace{2mm}~
    Serialization methods can significantly influence loan approval or denial, which, in turn, may have long-term consequences for individuals wrongly denied loans. 
    
\end{tcolorbox}
\subsection{Does Serialization Using Natural Language Texts Improve Performance?}
We hypothesized that using more natural input text would improve model performance, which motivated our inclusion of the \lift and \great serialization method (see Table~\ref{example_serialisation}). \lift
produced the best results for \llamasit on the Ghana dataset. However, this improvement did not hold consistently across all models and datasets, indicating that while natural language formats can be beneficial, their effectiveness is context-dependent.


\begin{tcolorbox}[
    colback=blue!3!white,
    colframe=blue!50!black,
    boxrule=0.5mm,       
    arc=2mm,             
    left=1mm,            
    right=1mm,           
    top=1mm,             
    bottom=1mm             
]
    {\fontsize{10}{20}\selectfont \faLightbulbO}\hspace{2mm}~Our results show that increasing the naturalness of input formatting does not consistently enhance model performance. 
\end{tcolorbox}

\subsection{Does Using Few-Shot Examples Improve the Decision-Making Abilities of LLMs?}

Given LLMs' subpar performance in the zero-shot experiments, we explored various methods to improve their decision-making capabilities through in-context learning(ICL). Figure~\ref{fig:shot_v_f1_per_data} presents the results from our ICL experiment, where we provide the model with varying numbers of n-shot examples, ranging from zero-shot $(n=0)$ to 8-shot across datasets and serialization formats. 
From Figure~\ref{fig:shot_v_f1_per_data}, providing more examples improves the loan approval task. Similarly, in Figure \ref{per_format_score_1x3_shared_legend}, we see average improvement with more examples. This is shown across all the serialization methods.
\begin{tcolorbox}[
    colback=blue!3!white,
    colframe=blue!50!black,
    boxrule=0.5mm,       
    arc=2mm,             
    left=1mm,            
    right=1mm,           
    top=1mm,             
]
    {\fontsize{10}{20}\selectfont \faLightbulbO}\hspace{2mm}~Model performance improves with more example shots, improving LLM decision-making for loan approval.
\end{tcolorbox}

\begin{table}[!ht]
\centering
\resizebox{\columnwidth}{!}{%
\begin{tabular}{lll|ll|ll}
\toprule
\textbf{Datasets:} & \multicolumn{2}{c}{\textbf{Germany}}  & \multicolumn{2}{c}{\textbf{Ghana}} & \multicolumn{2}{c}{\textbf{U.S.}}  \\
\textbf{Fairness Metrics:}& \textbf{SP} & \textbf{EO}    & \textbf{SP} & \textbf{EO}   & \textbf{SP} & \textbf{EO}    \\
\midrule
\multicolumn{4}{l}{\textit{\textbf{Baseline models }}} \\
        \textit{Zero} & 0.00 & 0.00  & 0.00 & 0.00  & 0.00 & 0.00 \\ 
        \textit{One}  & 0.00 & 0.00  & 0.00 & 0.00 &  0.00 & 0.00 \\
        \textit{Random} &  0.15 & 0.38 &  0.02& -0.03  & 0.04 & 0.35 \\
\midrule\multicolumn{4}{l}{\textit{\textbf{Benchmark model}}} \\
        \textit{Logistic Regression} &  -0.03 & -0.08 &  -0.04 & 0.05  & -0.02 & -0.01 \\ 
\midrule
\multicolumn{4}{l}{\textit{\textbf{Models Fine-tuned for Finance}}} \\
        \finmafull  & \textbf{\textcolor{red}{0.13}} & \textbf{\textcolor{red}{0.16}}  & 0.03 & \textbf{\textcolor{red}{0.06}}  & 0.00 & 0.00 \\ 
        \finmanlp & 0.07 & 0.07  & 0.00 & 0.01 & 0.00 & 0.00 \\ 
\midrule
\multicolumn{4}{l}{\textit{\textbf{Mid range open-source  base models}}} \\
      \llamathree &  0.00 & 0.00  & 0.00 & 0.00  & 0.00 & 0.00 \\ 
        \gemmanine  & 0.05 & 0.05  & -0.03 & -0.04 &  \textbf{\textcolor{red}{-0.06}} & -0.11 \\
\midrule
\multicolumn{4}{l}{\textit{\textbf{Mid range open-source  instruction tuned models}}} \\
        \llamathreeit &  0.03 & 0.06  & 0.00 & 0.00  & 0.01 & 0.02 \\ 
        \gemmanineit  & 0.01 & 0.01  & 0.03 & 0.04  & -0.04 & 0.13 \\ 
        \midrule
\multicolumn{4}{l}{\textit{\textbf{Large range open-source  instruction tuned models}}} \\
        \llamasit  & -0.03 & 0.01  & 0.00 & 0.00  & -0.01 & 0.03 \\ 
        \gemmatsit  & -0.01 & -0.02  & 0.00 & 0.02  & 0.04 & \textbf{\textcolor{red}{0.17}} \\
        \midrule
\multicolumn{4}{l}{\textit{\textbf{Large range open-source  base models}}} \\
        \llamas  & -0.05 & -0.05 & 0.00 & 0.00  & 0.00 & 0.00 \\ 
        \gemmats  & 0.00 & 0.03  & 0.00 & -0.02 & 0.01 & 0.07 \\ \bottomrule
        
    \end{tabular}
}

    \caption{\textbf{Zero-shot fairness metrics across regions for \json serialization.} 
    The red colour shows high bias across comparing models, excluding baselines. }
\label{tab:zeroshottable-fairness}
\end{table}

\begin{figure*}[!h]
\centering
\includegraphics[width=\linewidth]{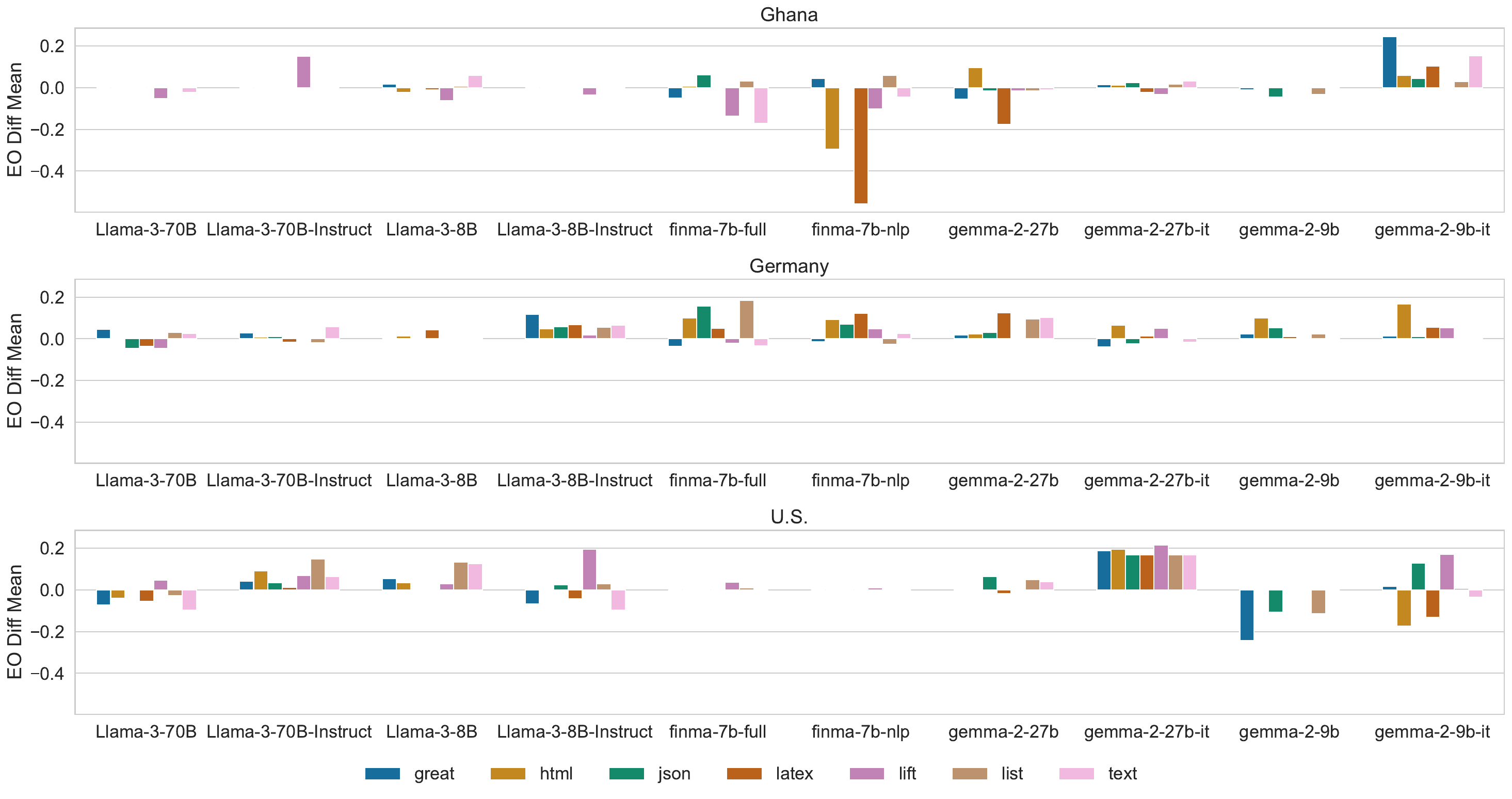}
\caption{\textbf{Mean difference in EO for different serialization methods and models.} Finance-based models show higher gender-based disparity for certain serializations, while the results are highly region and format-dependent.} 
\label{fig:eo_diff_mean}
\end{figure*}

\subsection{
How Does Model Fairness Vary Across Datasets?}
Baseline models from Table~\ref{tab:zeroshottable-fairness} all show no discrimination in terms of equality of opportunity (EO) and statistical parity (SP) except the \random.
However, we see high discrimination in terms of both EO and SP with the \finmafull for the Germany dataset. Similarly, we see this model also returns the highest disparity in terms of EO in the Ghana dataset. It is interesting to note that this model, among the other models selected in this study, is the only one fine-tuned for finance. This, therefore, opens up interesting research directions for further investigating the fairness of downstream tasks that have been trained with this model.  In a similar light, \gemmatsit returns the highest disparity in terms of EO for the U.S. dataset. 
On the contrary, \llamathree has no disparity in terms of both fairness metrics on the Germany data. Further highlighting that different models penalize sensitive groups differently.
Additionally, examining fairness by conducting few-shot experiments showed that few-shot examples (e.g., $n = 8$) can introduce significant fairness disparities in Equality of Opportunity (EO), with differences exceeding $0.10$ for certain serialization methods in the Ghana dataset (see Figure \ref{fig:eo_diff_mean}).





\begin{tcolorbox}[
    colback=blue!3!white,
    colframe=blue!50!black,
    boxrule=0.5mm,       
    arc=1mm,             
    left=1mm,            
    right=1mm,           
    top=1mm,             
    bottom=0mm  
]
    {\fontsize{10}{20}\selectfont \faLightbulbO}\hspace{1mm}~
      LLMs fine-tuned on financial datasets have the potential to amplify existing historical gender bias.
\end{tcolorbox}


\vspace{-0.25cm}
\subsection{What Is the Fairness F1 Score Tradeoffs?}
 Following the best-performing models, as shown in Figure~\ref{per_format_score_1x3_shared_legend}, we assess the fairness of these models in Figure~\ref{fig:eo_diff_mean}. The \gemmatsit model shows a degree of disparity for the U.S. data. In the case of the best-performing model for Germany, \llamasit does not show a higher level of unfairness compared to the \llamasit and \finmafull models. The \gemmanineit model shows a higher disparity in the EO difference. The \finmafull model shows a higher disparity in terms of EO in both the Ghana and Germany datasets. The negative EO difference highlights that the model discriminates against the non-protected group, which in this case is males.

\begin{tcolorbox}[
    colback=blue!3!white,
    colframe=blue!50!black,
    boxrule=0.5mm,       
    arc=2mm,             
    left=1mm,            
    right=1mm,           
    top=1mm,             
    bottom=1mm              
]
    {\fontsize{10}{20}\selectfont \faLightbulbO}\hspace{2mm}~ Financial-based models exhibit greater disparities in EO mean difference and high performance does not equate to fairness.
\end{tcolorbox}

\subsection{How Does Prompt Sensitivity Vary Across Different Regions and Models?}

The results in Figure~\ref{fig:zero-shot-format-serialization} represent the average performance across three different prompts, with error bars indicating the sensitivity to prompt variations. We observe relatively low prompt sensitivity in the U.S. and Ghana datasets, whereas the German dataset exhibits significantly higher sensitivity to prompt differences. 


\begin{tcolorbox}[
    colback=blue!3!white,
    colframe=blue!50!black,
    boxrule=0.5mm,       
    arc=2mm,             
    left=1mm,            
    right=1mm,           
    top=1mm,             
    bottom=1mm              
]
    {\fontsize{10}{20}\selectfont \faLightbulbO}\hspace{2mm}~LLM performance sensitivity to prompts varies across data sources—some datasets exhibit stable results across prompts, while others show significant variability.
\end{tcolorbox}

\subsection{How Does Model Size Relate to Performance and Fairness?}

We assess the effect of model scale by evaluating multiple sizes of both the LLaMA and Gemma families in Figure \ref{fig:zero-shot-format-serialization}. Across LLaMA variants, expanding parameter counts yields only marginal performance changes. In contrast, Gemma exhibits pronounced performance gains as size increases, a pattern that re-emerges in the fairness analysis (Figure \ref{fig:eo_diff_mean}), where the Gemma models’ equality-of-opportunity scores are highly sensitive to model scale.

\subsection{Does Instruction Tuning Affect Model Performance and Fairness Scores?}

We further investigated the impact of instruction tuning on accuracy and fairness by comparing the base and instruction-tuned variants of the LLaMA and Gemma families (Figure \ref{fig:zero-shot-format-serialization}). Instruction tuning has little effect on LLaMA, but its influence on Gemma depends on model size: the 9 B version loses accuracy. These shifts are also shaped by the choice of serialization. For example, the instruction-tuned Gemma improves fairness on the United States dataset in some formats, yet becomes more biased in others. 


\subsection{Do Few-Shot Examples Improve Fairness?}

In the German dataset, 
with reference to 
Figure~\ref{icl-fair},
Few-shot examples (e.g., $n=8$) can lead to significant fairness disparities in equality of opportunity (EO), reaching differences of over $0.10$ for some serialization methods in the Ghana datasets. The U.S. dataset shows greater sensitivity to few-shot examples, with models exhibiting a decline in fairness scores.


\begin{tcolorbox}[
    colback=blue!3!white,
    colframe=blue!50!black,
    boxrule=0.5mm,       
    arc=2mm,             
    left=1mm,            
    right=1mm,           
    top=1mm,             
    bottom=1mm             
]
    {\fontsize{10}{20}\selectfont \faLightbulbO}\hspace{2mm}~Fairness in few-shot learning is highly context-dependent. While more examples can sometimes reduce disparities, the impact is not universal, underscoring the importance of carefully selecting and evaluating serialization methods to ensure fairness.
\end{tcolorbox}

\section{Discussion and Conclusion}
\label{conclusion}


\paragraph{Summary.} The ability of LLMs to handle structured tabular data for high-stakes tasks like loan approvals remains under-explored. This work evaluates how different serialization methods (\json, \lift, \Text) and in-context learning (ICL) impact the fairness and accuracy of LLMs across diverse regional datasets (Ghana, Germany, United States). We find that, in zero-shot scenarios, all LLMs perform worse than a \logistic baseline, frequently defaulting to uniform approval or denial. Modest improvements only emerge with a few in-context examples, largely influenced by serialization format and dataset rather than model.

\paragraph{Fairness implications of LLMs in finance.}
The results indicate that LLMs fine-tuned on financial datasets cannot yet be fully trusted for high-stakes financial decisions. Therefore, careful attention to data representation is at least as critical as the choice of model. To further address fairness concerns, employing more balanced datasets and ensuring a transparent decision-making process could be beneficial. This transparency is particularly important, as prior decisions made by banks can significantly impact the long-term creditworthiness of applicants~\cite{majumdar2025causal}.

\paragraph{Recommendation for practitioners.}
We recommend that practitioners retain thoroughly validated tabular models as a baseline and treat LLM outputs only as decision support until they demonstrably exceed that baseline in both accuracy and fairness. During and after deployment, models should be stress-tested on multiple serialization approaches and on regionally diverse datasets to ensure robustness. Benchmarking must extend beyond raw performance scores to include a suite of fairness and accuracy metrics so that improvements in prediction quality do not mask emerging biases.

\paragraph{Future work.} Explore serialization-robust training, fairness-aware optimization, interpretability methods that expose feature reliance, and broader multilingual datasets that capture diverse regions.

\section*{Limitations}
\label{sec:limitation}

\paragraph{Dataset Differences.}
In our work, we examined data sources from different regions, but a detailed study and analysis of the differences between these datasets is crucial. We used the default column names and values for all datasets. However, some of our serialization methods, such as \lift, aimed to improve column names by correcting spelling errors and related mistakes inherent in the datasets. We acknowledge that there may still be variances that have not been captured and need further investigation.

\paragraph{More Datasets.}

This study focused on three datasets from distinct geographical regions. While incorporating additional datasets with greater variability could improve the research, we maintained this scope to align with the study’s objectives and constraints.



\paragraph{LLMs Covered in the Work.}

This work covers a limited number of LLMs, and we mostly focused on models that we believed, to the best of our knowledge, would be adapted to several use cases because of popularity, open source and continued support by organizations that release them. We purposefully left our closed-sourced model due to resource constraints and limited flexibility for experimentation, particularly around fine-grained control of inputs and internal mechanisms.




\paragraph{Prompt Design.}
In this study, we generated prompts by referencing similar research works. While certain prompt structures may outperform others, a comprehensive exploration of prompt engineering techniques is beyond this work’s scope due to the extensive number of experiments conducted. We acknowledge the importance of this aspect and propose it as a direction for future research.


\paragraph{Explaining Model Behaviour.} 
We conducted token attribution experiments to better understand the reasoning behind model behaviour. However, as the results were inconclusive, we have not included a detailed discussion in the main text. Instead, a comprehensive account of the findings can be found in Appendix~\ref{token-attrbution-experiments}.

 
\section*{Acknowledgment}

\emph{Israel Abebe Azime} would like to thank the German Federal Ministry of Education and Research and the German federal states (http://www.nhr-verein.de/en/our-partners) for supporting this work/project as part of the National High-Performance Computing (NHR) joint funding program.  \emph{Deborah D. Kanubala} and \emph{Isabel Valera} are supported by the European Union (ERC-2021-STG, SAML, 101040177). \emph{Tejumade Afonja} is partially supported by ELSA – European Lighthouse on Secure and Safe AI funded by the European Union under grant agreement No. 101070617.\\
Views and opinions expressed are, however, those of the author(s) only and do not necessarily reflect those of the funding organizations and neither can they be held responsible for them.

\bibliography{references}
\newpage

\newpage
\onecolumn
\appendix

\section*{Appendix}
\label{appendix}

\section{Metrics}
\label{metrics}

In evaluating the performance of Large Language Models (LLMs), we employ several key metrics to assess their predictive accuracy. These metrics provide a comprehensive view of how well the models align with ground truth labels. 
\vspace{2mm}
\vspace{2mm}
\begin{definition}[Weighted-Average F1 Score:] The weighted average F1 score calculates the F1 score for each class independently and then combines them using weights that are proportional to the number of true labels in each class.

$$ \textnormal{Weighted-Average F1 Score}= \sum_{i=1}^C w_i \times \textnormal{F1 Score}_i
$$
where $$w_i = \frac{\textnormal{No. of samples in class } i}{\textnormal{Total number of samples}}$$
and $C$ is the number of classes in the dataset.
\end{definition}

\vspace{1cm}


\section{Model Evaluation Setup}
\label{eval-setup}

For this task, we utilized EleutherAI’s open-source Language Model Evaluation Harness (lm-eval) framework \cite{eval-harness}. We created custom configurations for each task and looked at log-likelihood prediction for each possible token and decided possible generation from the possible class outputs. we created 3 different prompts for each data sources and evaluated on same generation settings. 


\FloatBarrier
\section{Dataset Description and Analysis}
\label{sec:data-descriptions}

Table~\ref{table:loanpredictions_features},~\ref{table:ghana_credit_features}, and \ref{table:germancredit_features} present the features included in the datasets. We use the target features as output classes, and for serializations that convert feature names to text, we correct spelling to improve clarity and expressiveness.

\begin{table*}[!h]
\centering
\resizebox{0.70\columnwidth}{!}{%
\begin{tabular}{|l|l|}
\hline
\textbf{Feature Name} & \textbf{Description} \\ \hline
Loan\_ID & Unique identifier for the loan \\ 
Gender & Gender of the applicant \\
Married & Marital status of the applicant \\ 
Dependents & Number of dependents of the applicant \\ 
Education & Education level of the applicant \\ 
Self\_Employed & Whether the applicant is self-employed \\ 
ApplicantIncome & Income of the applicant \\ 
CoapplicantIncome & Income of the co-applicant \\ 
LoanAmount & Loan amount requested \\ 
Loan\_Amount\_Term & Term of the loan in months \\ 
Credit\_History & Credit history of the applicant \\ 
Property\_Area & Area type of the property \\ \hline
Loan\_Status & Status of the loan (e.g., Loan paid or not ) \\ \bottomrule
\end{tabular}
}
\caption{Description of Features for US Loan Predictions Dataset}
\label{table:loanpredictions_features}
\end{table*}

\begin{table*}[!h]
\centering
\resizebox{0.70\columnwidth}{!}{%
\begin{tabular}{|l|l|}
\hline
\textbf{Feature Name} & \textbf{Description} \\ \hline
sex & Gender of the applicant \\ 
amnt req & Amount requested for the loan \\
ration & Ratio of the amount granted to the amount requested \\ 
maturity & Maturity period of the loan \\ 
assets val & Value of the applicant's assets \\ 
dec profit & Decision on the profit potential \\ 
xperience & Experience of the applicant \\ 
educatn & Education level of the applicant \\ 
age & Age of the applicant \\ 
collateral & Collateral provided for the loan \\ 
locatn & Location of the applicant \\ 
guarantor & Guarantor for the loan \\ 
relatnshp & Relationship with the financial institution \\ 
purpose & Purpose of the loan \\ 
sector & Economic sector of the applicant \\ 
savings & Savings of the applicant \\ \hline
target & Loan amount requested granted or not \\ \hline
\end{tabular}
}
\caption{Description of Features for Ghana Credit Rationing Dataset}
\label{table:ghana_credit_features}
\end{table*}

\label{data-description}
\begin{table*}[!h]
\centering
\resizebox{0.90\columnwidth}{!}{%
\begin{tabular}{|l|l|}
\hline
\textbf{Feature Name} & \textbf{Description} \\ \hline
gender & The gender of the individual \\ 
checking\_status & The status of the individual's checking account \\ 
duration & Duration of the credit in months \\ 
credit\_history & Credit history of the individual \\ 
purpose & Purpose of the credit \\ 
credit\_amount & Amount of credit requested \\ 
savings\_status & Status of the individual's savings account \\ 
employment & Employment status of the individual \\ 
installment\_commitment & Installment commitment as a percentage of disposable income \\ 
other\_parties & Other parties related to the credit \\ 
residence\_since & Number of years the individual has lived in their current residence \\ 
property\_magnitude & Value or magnitude of property \\ 
age & Age of the individual \\ 
other\_payment\_plans & Other payment plans that the individual has \\ 
housing & Housing status of the individual \\ 
existing\_credits & Number of existing credits at this bank \\ 
job & Job status of the individual \\ 
num\_dependents & Number of dependents \\ 
own\_telephone & Whether the individual owns a telephone \\ 
foreign\_worker & Whether the individual is a foreign worker \\ \hline
class & Classification of the credit (e.g., good or bad) \\ \bottomrule
\end{tabular}
}
\caption{Description of Features in German Credit Dataset}
\label{table:germancredit_features}
\end{table*}

\begin{figure*}[!h]
\centering
\includegraphics[width=0.75\linewidth]{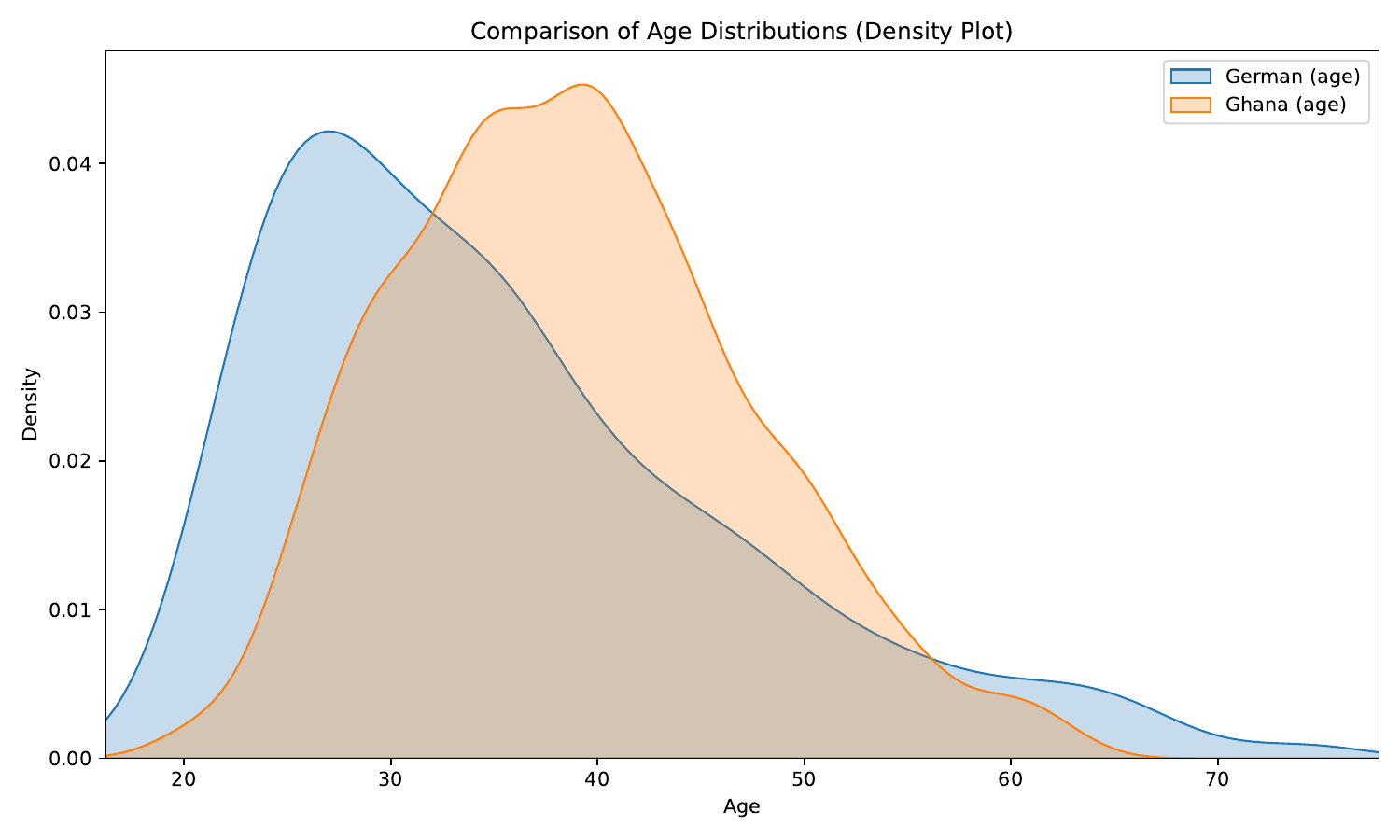}
\includegraphics[width=0.75\linewidth]{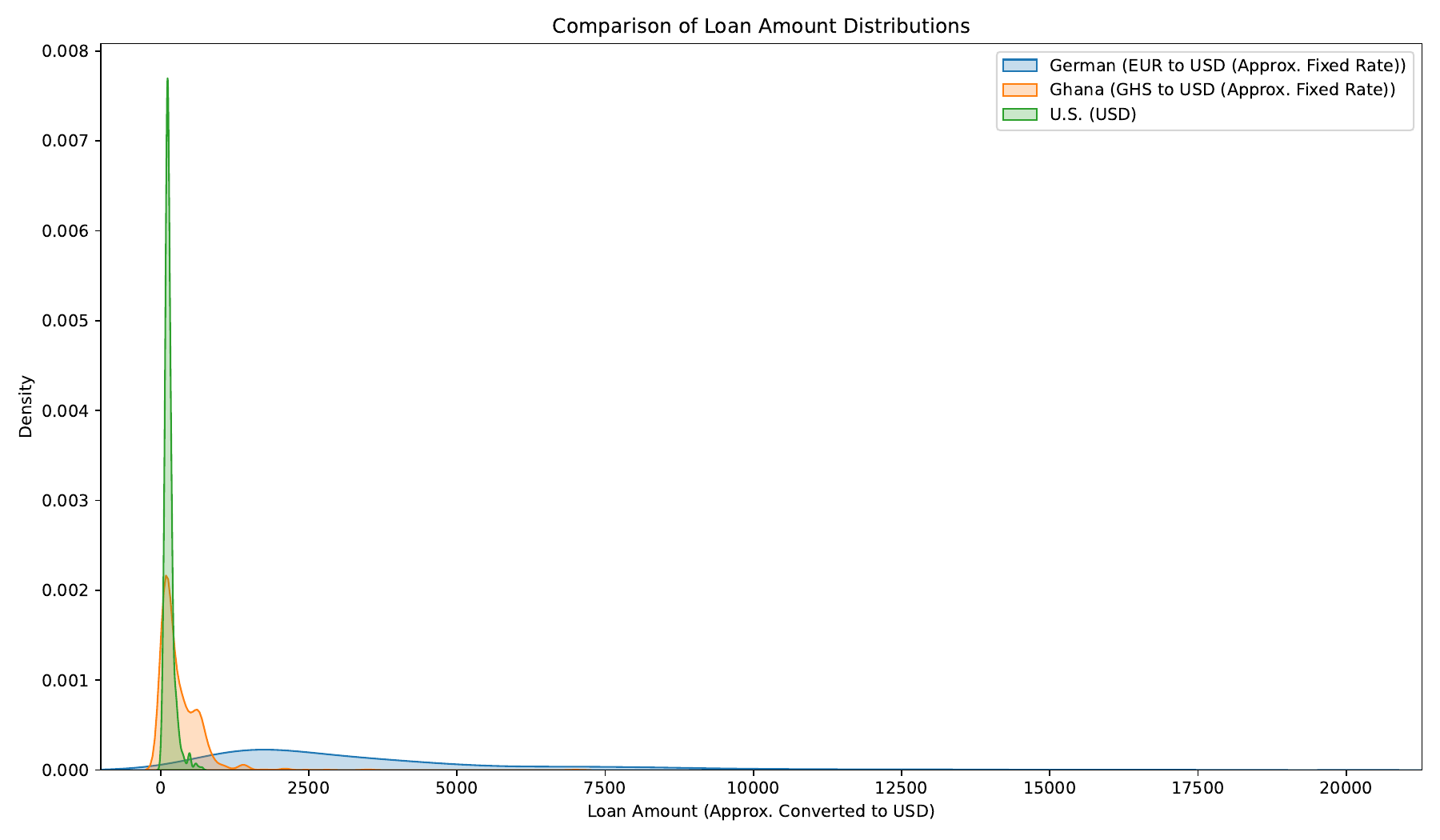}
\caption{KDE plot comparing age and loan amount distributions across datasets, highlighting inherent socio-economic and cultural disparities. The age distribution reveals that the Ghana dataset skews older, with a concentration in the 30-50 age range, while the German dataset shows a relatively younger distribution peaking around the 20-30 age range. Loan amounts are predominantly smaller in both Ghana and U.S. datasets, with the German dataset exhibiting a broader distribution range, indicating socio-economic and lending disparities across regions.}
\label{}
\end{figure*}

\FloatBarrier
\section{Serialization}
\label{app:serialization}

Table~\ref{example_serialisation_full} shows examples of the six (6) different serialization methods employed in this work. We considered straightforward default values, such as \json and \List, to more structured and natural language text-like formats, such as \html, \latex, \Text~\citep{hegselmann2023tabllm}, \great~\citep{borisov2022language} and \lift~\citep{dinh2022lift}.

\begin{table*}[!h]
\footnotesize
\centering
\begin{tabular}{l|p{10cm}}
\toprule
\textbf{Serialization}& \textbf{Example Template} \\
\midrule
\texttt{JSON (default)} & 

 \{age: 32, sex: female, loan duration: 48 months,
 purpose: education\} 
  \\
\midrule
 \texttt{List}& \texttt{- age: 32 \newline - sex: female \newline - loan duration: 48 months \newline - purpose: education} \\
 \midrule
\texttt{GReaT \cite{borisov2022language}} & \texttt{age is 32, sex is female, loan duration is 48 months, loan purpose is education} \\
\midrule
 \texttt{Text} & \texttt{The age is 32. The sex is female. The loan duration is 48 months. The purpose is education.}\\
\midrule
\texttt{LIFT \cite{dinh2022lift}}& \texttt{A 32-year-old female is applying for a loan for 48 months for education purposes.}\\
\midrule         
\texttt{HTML}& \texttt{<table><thead>\newline <tr><th>age</th> <th>sex</th>
         \newline . . . 
           \newline<tr><td>32</td><td>female</td>
           \newline . . . 
           \newline
           </tr> \newline</tbody></table>} \\
\midrule
\texttt{Latex} &\begin{verbatim}
\begin{tabular}{lrrr} 
\toprule 
age & sex & loan duration & purpuse  \\
\midrule 
32 & female & 48 month & education \\
\end{tabular}
\end{verbatim}
\\
\bottomrule 
\end{tabular}
\caption{\textbf{Comparison of serialization formats for loan applicant information.} This table presents example templates for representing loan applicant data with four features (age and sex, loan duration and purpose). JSON is assumed as the default format. The selected serialization formats ensure diverse data representation, balancing availability across different formats, naturalness, and alignment with prior work. \label{example_serialisation_full}
}
\end{table*}

\begin{figure*}[!ht]
\centering
\includegraphics[width=0.8\textwidth]{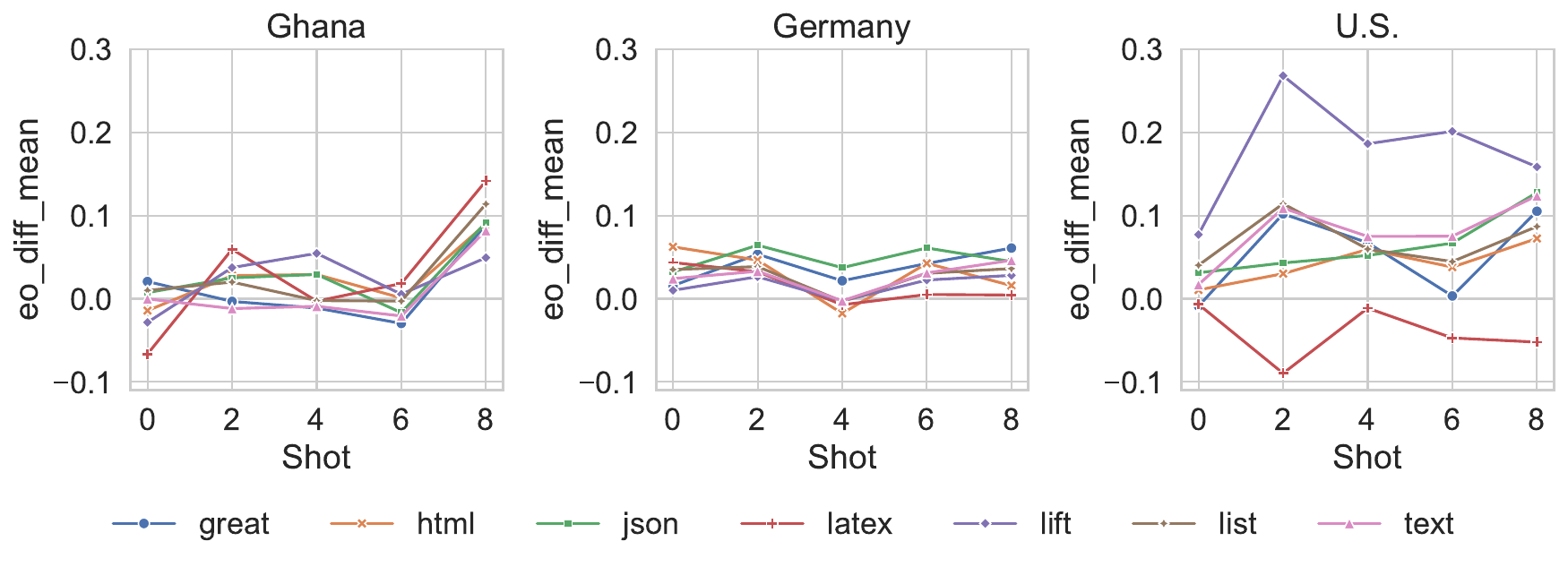}
\caption{\textbf{Equality of Opportunity Difference for Few-Shot Learning Across Serialization Methods and Datasets.}
In-context learning (ICL) does not consistently reduce bias; in some cases, models exhibit significantly unfair behavior, particularly in certain shot configurations.} 
\label{icl-fair}
\end{figure*}

\FloatBarrier
\section{More Fairness Scores}
\label{fewshot-fairness}

\begin{figure*}[!h]
\centering
\includegraphics[width=\textwidth]{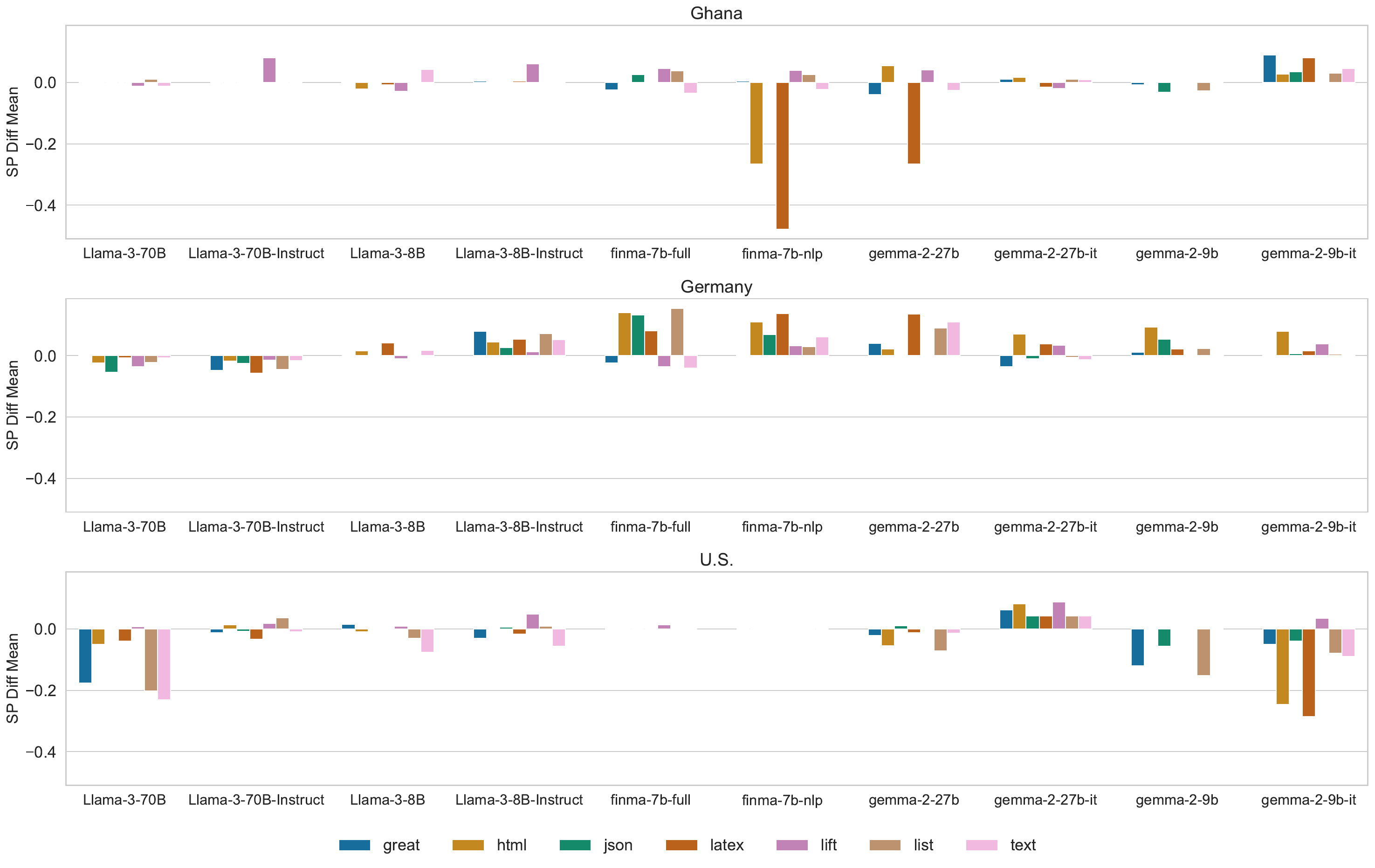}
\caption{\textbf{Statistical Parity Difference Across Serialization Methods and Models} This figure illustrates the statistical parity difference for various serialization methods and models. We observe that financial models exhibit notably high bias, particularly for the Ghana and Germany datasets.}
\label{fig:sp_fair}
\end{figure*}

We investigate additional questions, particularly the relationship between fairness scores and In-Context Learning (ICL) performance. Specifically, we analyze how variations in fairness scores impact ICL results, as illustrated in Figure \ref{icl-fair}. In Figure \ref{fig:sp_fair}, we present the statistical parity difference across various serialization methods and models. This analysis aims to examine how different serialization techniques impact fairness, providing insights into potential biases introduced by these encoding strategies.. This exploration aims to provide deeper insights into potential biases and the extent to which fairness considerations influence model performance in different settings.

\FloatBarrier
\section{Prompt Examples}
In Figure \ref{tab:prompt_templates}, we employed straightforward and minimally complex prompts for the task to maintain simplicity and consistency. For each task, we carefully adapted the prompt while ensuring alignment with the specific requirements of the evaluation. However, we intentionally chose not to modify the output classes or introduce entirely new prompts across different tasks, as doing so could have introduced unintended variables that might influence the evaluation outcomes.

\begin{table*}[!ht]
\begin{center}
\scalebox{0.7}{
\begin{tabular}{p{210mm}}
\toprule
\textbf{Example of Loanpred Prompt} \\
\midrule
\texttt{You are an expert LLM with extensive knowledge in financial analysis. Your task is to evaluate a person's profile and determine their likelihood of repaying a loan. 
Respond with 'Y' if the person is likely to repay the loan, and 'N' if they are not likely to repay.} \\ 
\\
\texttt{Profile: \{profile\}} \\      
\\
\texttt{Answer:}  \\ 
\midrule
\texttt{You are a financial analysis expert with extensive knowledge in credit assessment. Review the individual’s profile to evaluate their likelihood of repaying the loan. Respond with 'Y' if repayment is likely, or 'N' if it is not.} \\ 
\\
\texttt{Profile: \{profile\}} \\      
\\
\texttt{Answer:}  \\ 
\midrule
\texttt{As a financial expert skilled in assessing creditworthiness, analyze the following profile and indicate the likelihood of loan repayment. Use 'Y' for likely repayment and 'N' for unlikely.} \\ 
\\
\texttt{Profile: \{profile\}} \\      
\\
\texttt{Answer:}  \\ 
\addlinespace
\midrule
\textbf{Example Of Germany Prompt} \\
\midrule
\texttt{You are an expert LLM with extensive knowledge in financial analysis. Your task is to evaluate a person's profile and determine their likelihood of repaying a loan.
                Respond with 'good' if the person is likely to repay the loan, and 'bad' if they are not likely to repay.} \\ 
                \\
\texttt{Profile: \{profile\}} \\   
\\
\texttt{Answer: }  \\ 
\midrule
\texttt{You are a financial assessment specialist with deep insights into creditworthiness. 
Review the profile below and indicate the repayment likelihood with 'good' if the individual is likely to repay the loan, or 'bad' if they are not.} \\ 
\\
\texttt{Profile: \{profile\}} \\      
\\
\texttt{Answer:}  \\ 
\midrule
\texttt{
Imagine you are a loan assessment expert with extensive experience in evaluating repayment potential. Analyze the details provided to judge whether repayment is probable.
 Use 'good' for likely repayment and 'bad' for unlikely.} \\ 
\\
\texttt{Profile: \{profile\}} \\      
\\
\texttt{Answer:}  \\ 
\midrule
\textbf{Example Of Ghana Prompt} \\
\midrule
\texttt{You are an expert LLM with extensive knowledge in financial analysis. Your task is to evaluate a person's profile and determine their likelihood of repaying a loan. Respond with 'Yes' if the person is likely to repay the loan, and 'No' if they are not likely to repay.} \\ 
                \\
\texttt{Profile: \{profile\}} \\   
\\
\texttt{Answer: }  \\ 
\midrule
\texttt{
You are a financial risk evaluator with expertise in creditworthiness. Review the individual’s profile and indicate their repayment likelihood. Use 'Yes' for likely repayment, or 'No' if repayment is unlikely.} \\ 
\\
\texttt{Profile: \{profile\}} \\      
\\
\texttt{Answer:}  \\ 
\midrule
\texttt{
As an expert in financial analysis, assess the following profile to determine the likelihood of loan repayment. Respond with 'Yes' if repayment is probable, and 'No' if it is not.} \\ 
\\
\texttt{Profile: \{profile\}} \\      
\\
\texttt{Answer:}  \\ 
\addlinespace
\bottomrule
\end{tabular}
}
\caption{\textbf{Example Prompts Used for the Task.} For each task, we created three distinct prompts, and the reported results represent the average performance across all three.} 
\label{tab:prompt_templates}
\end{center}
\end{table*}

\FloatBarrier
\section{In-Context Learning (ICL)}
\begin{figure*}[!h]
\centering
\includegraphics[width=4.0cm]{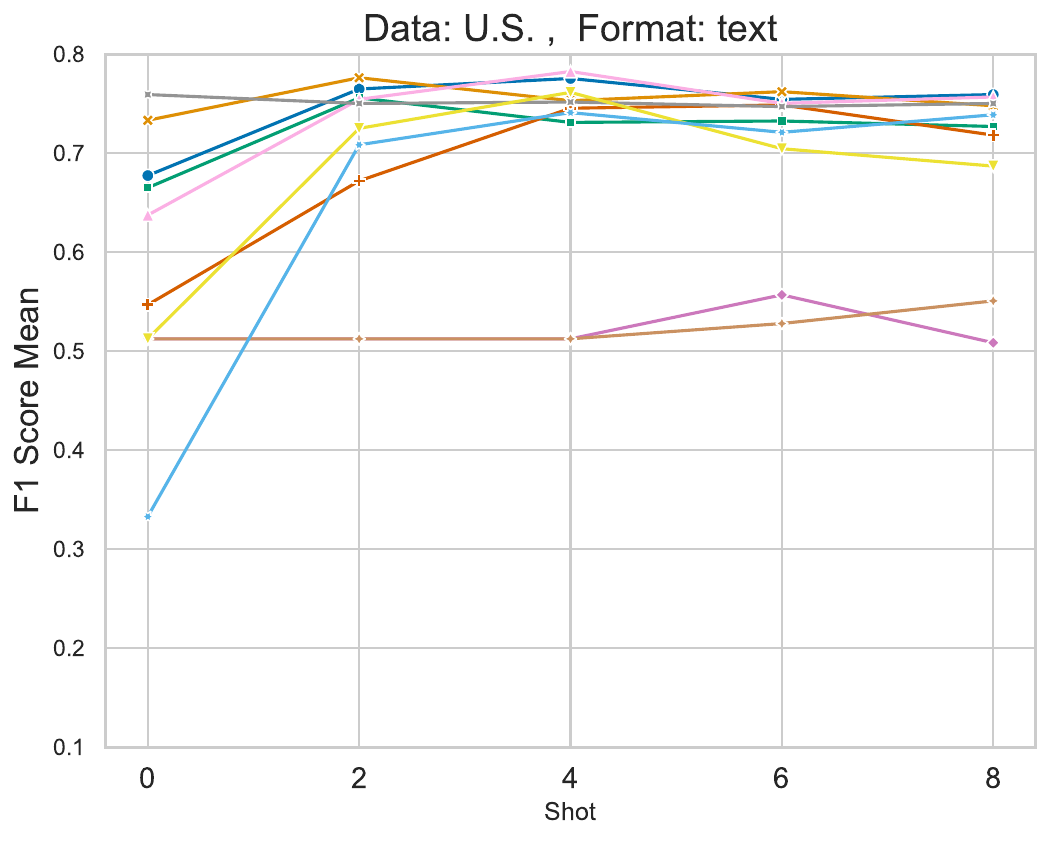}
\includegraphics[width=4.0cm]{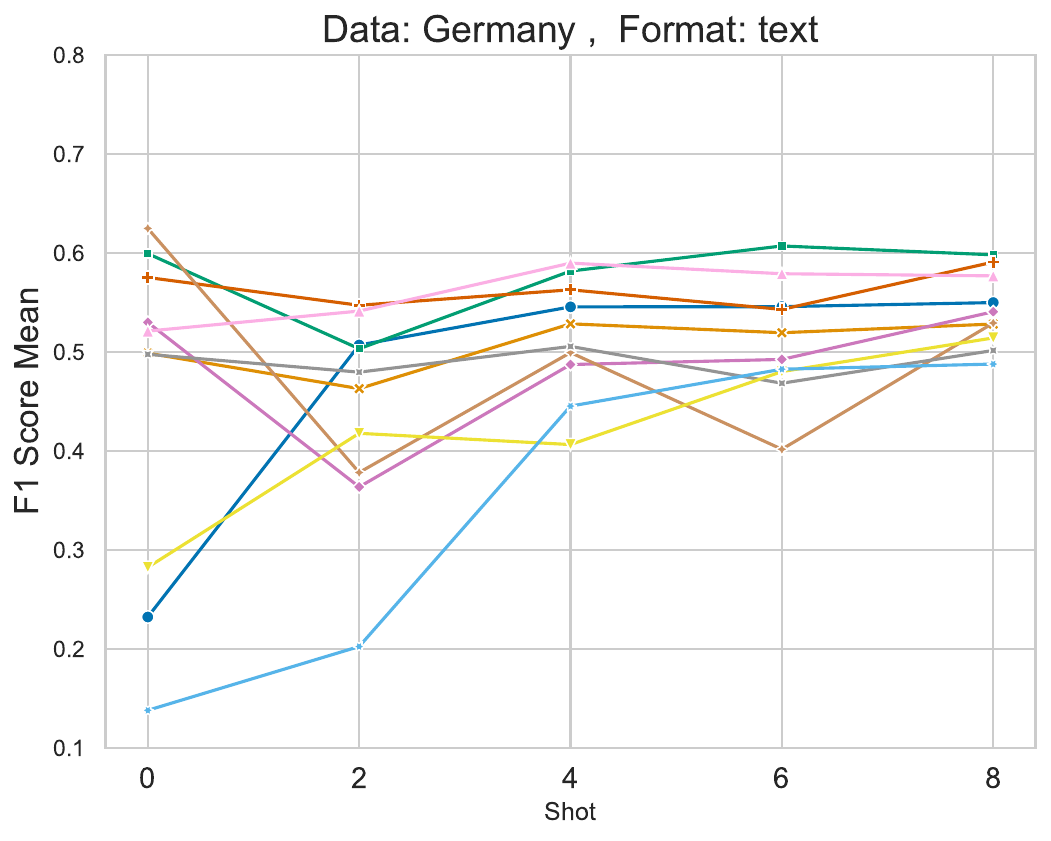}
\includegraphics[width=4.0cm]{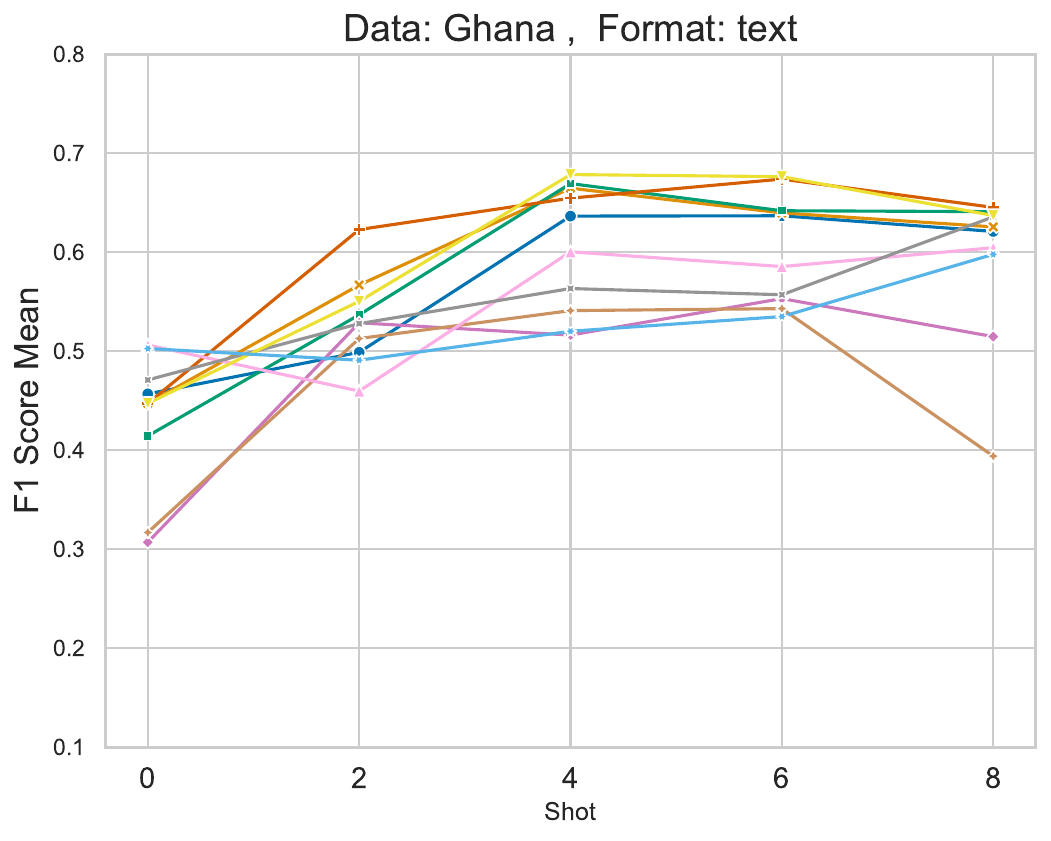}

\includegraphics[width=4.0cm]{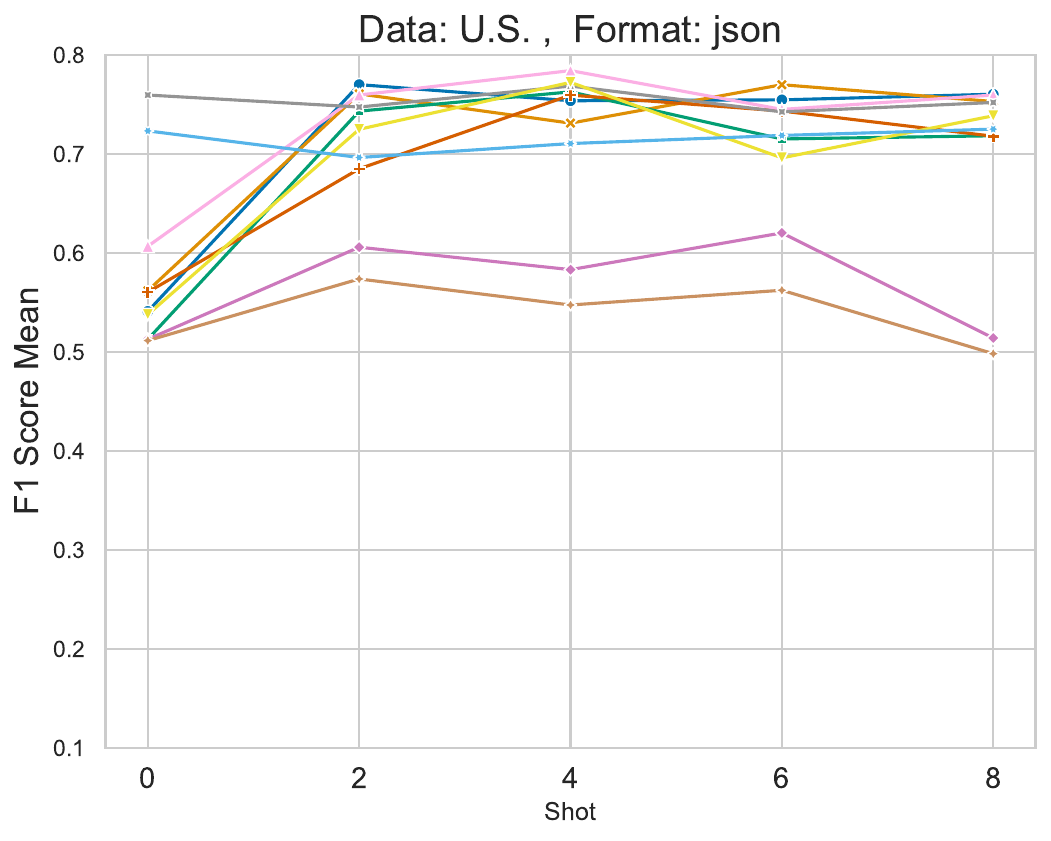}
\includegraphics[width=4.0cm]{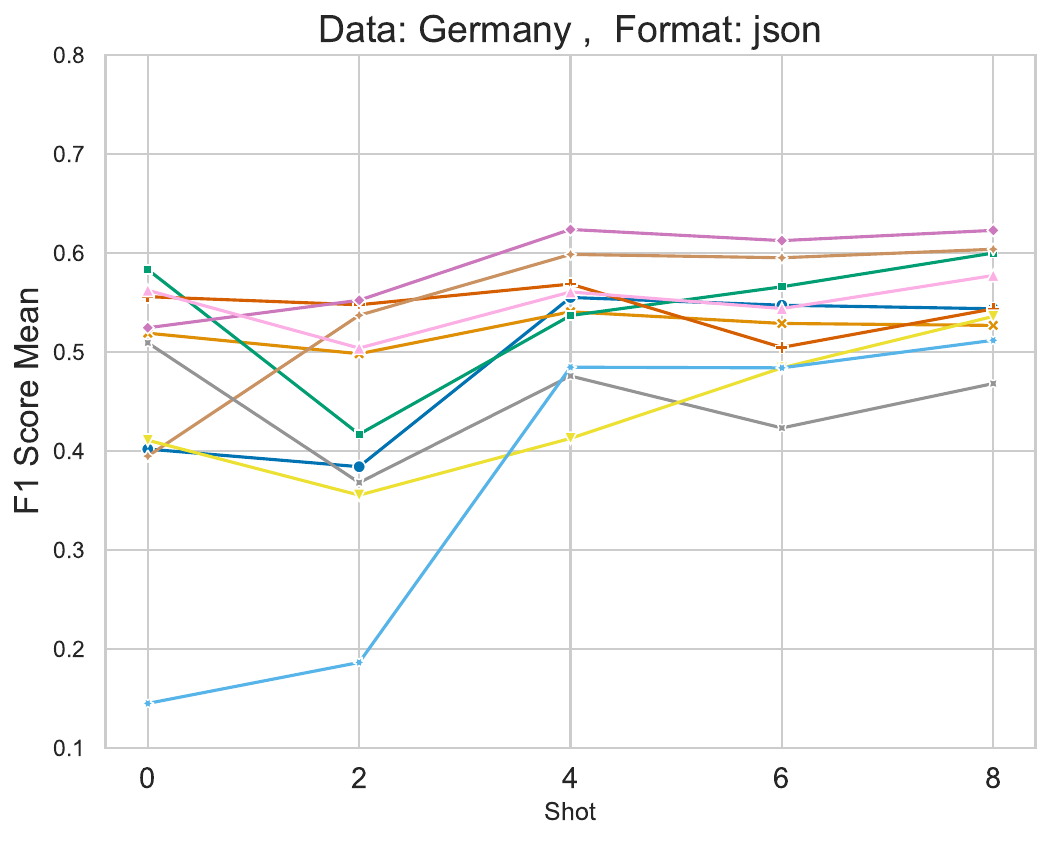}
\includegraphics[width=4.0cm]{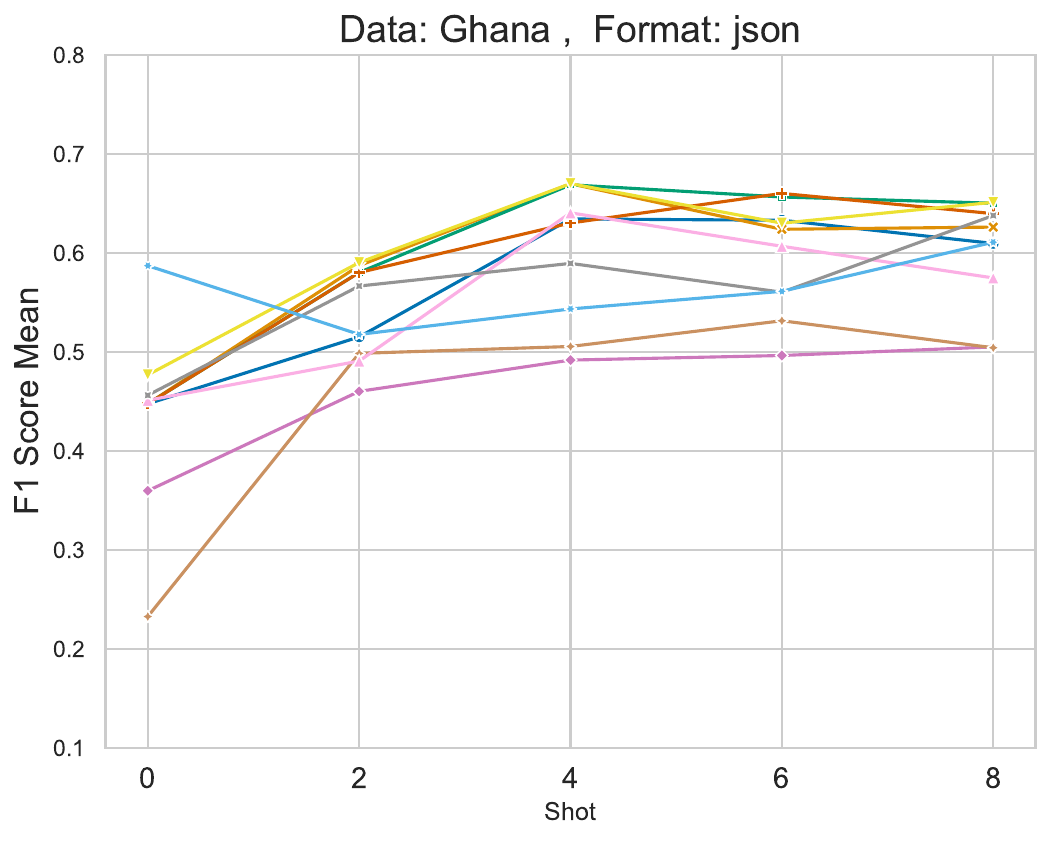}

\includegraphics[width=4.0cm]{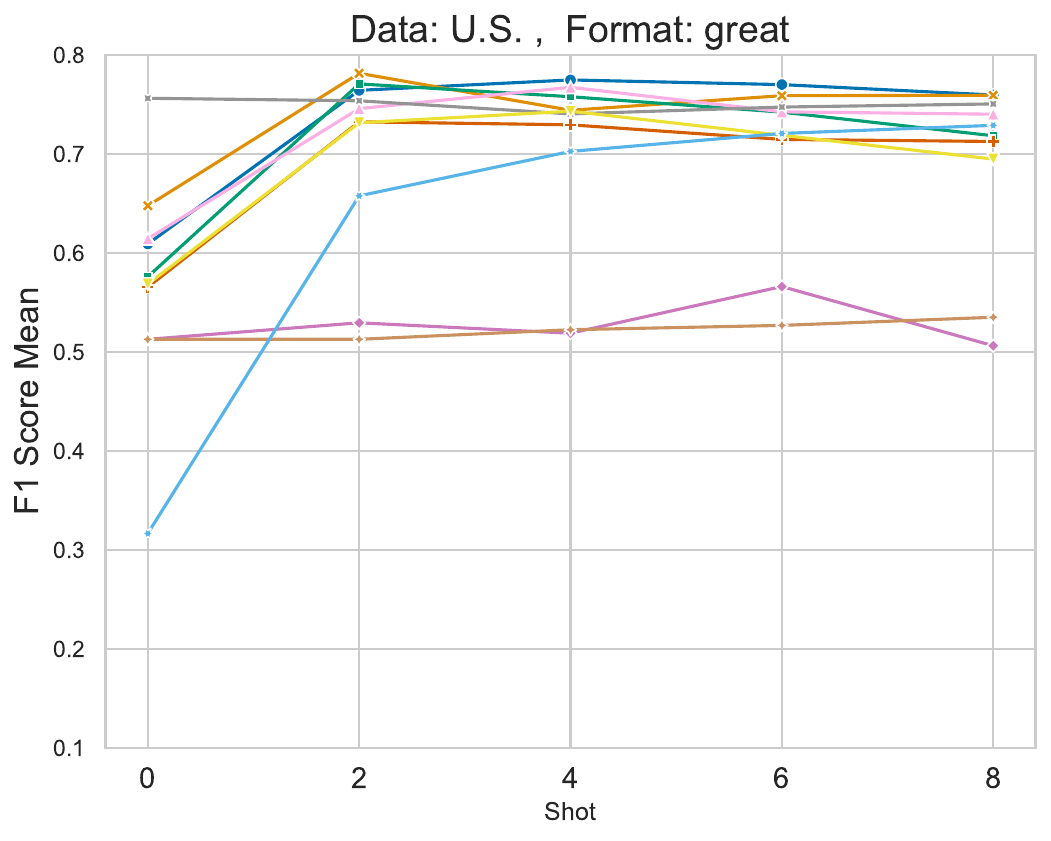}
\includegraphics[width=4.0cm]{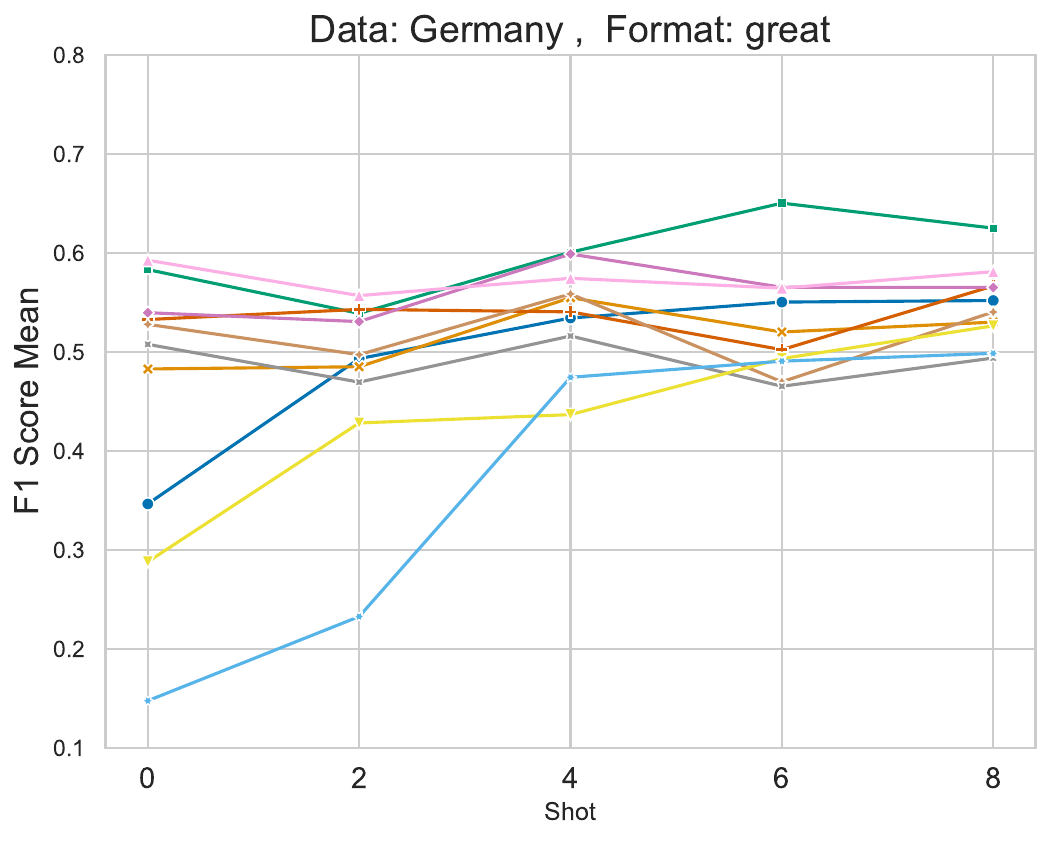}
\includegraphics[width=4.0cm]{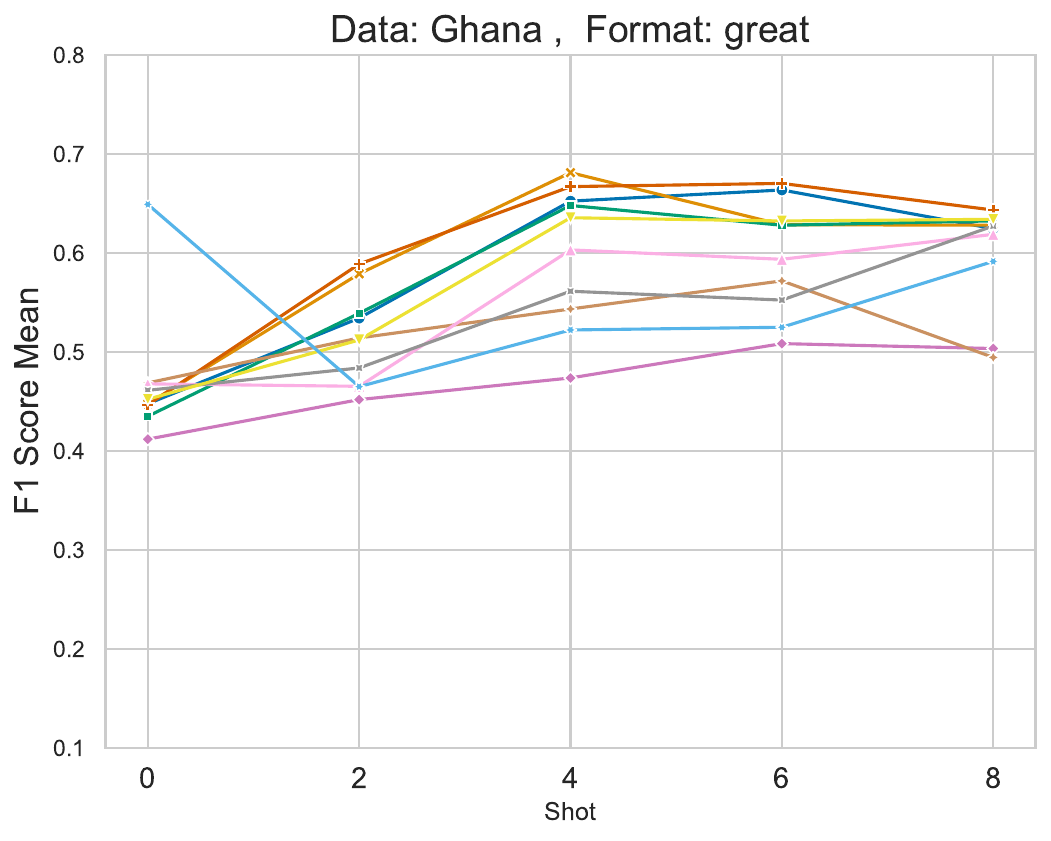}

\includegraphics[width=4.0cm]{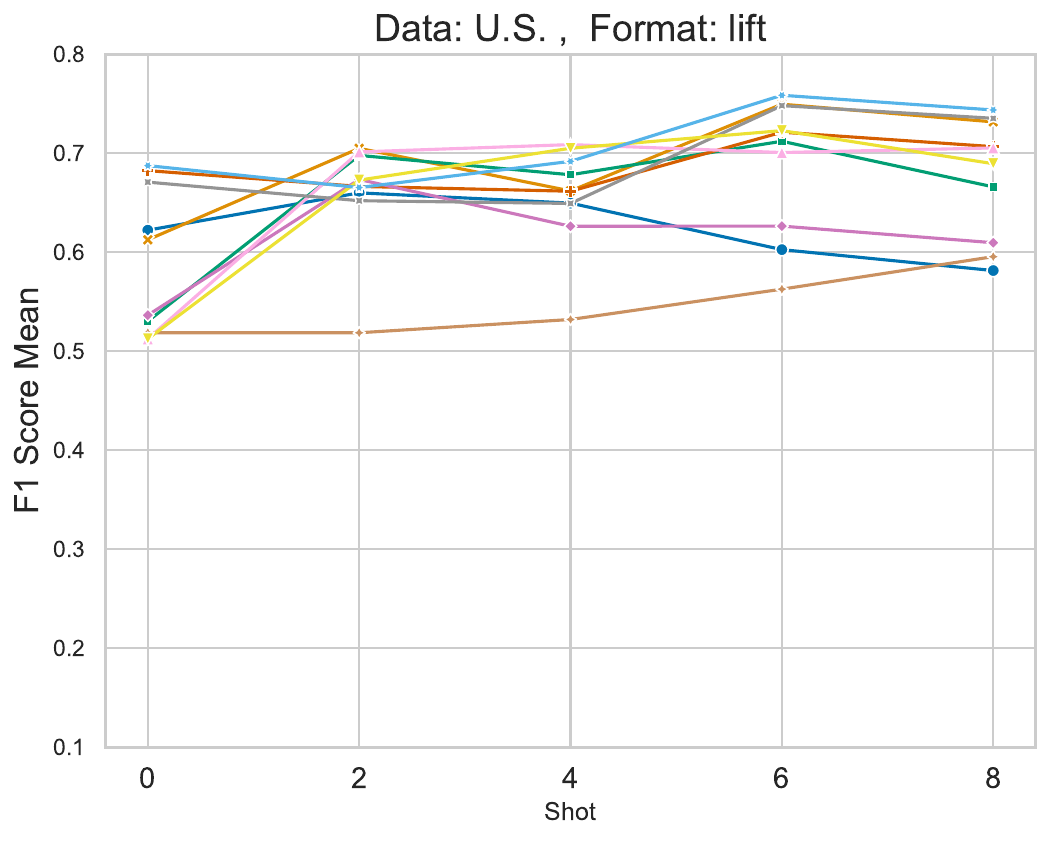}
\includegraphics[width=4.0cm]{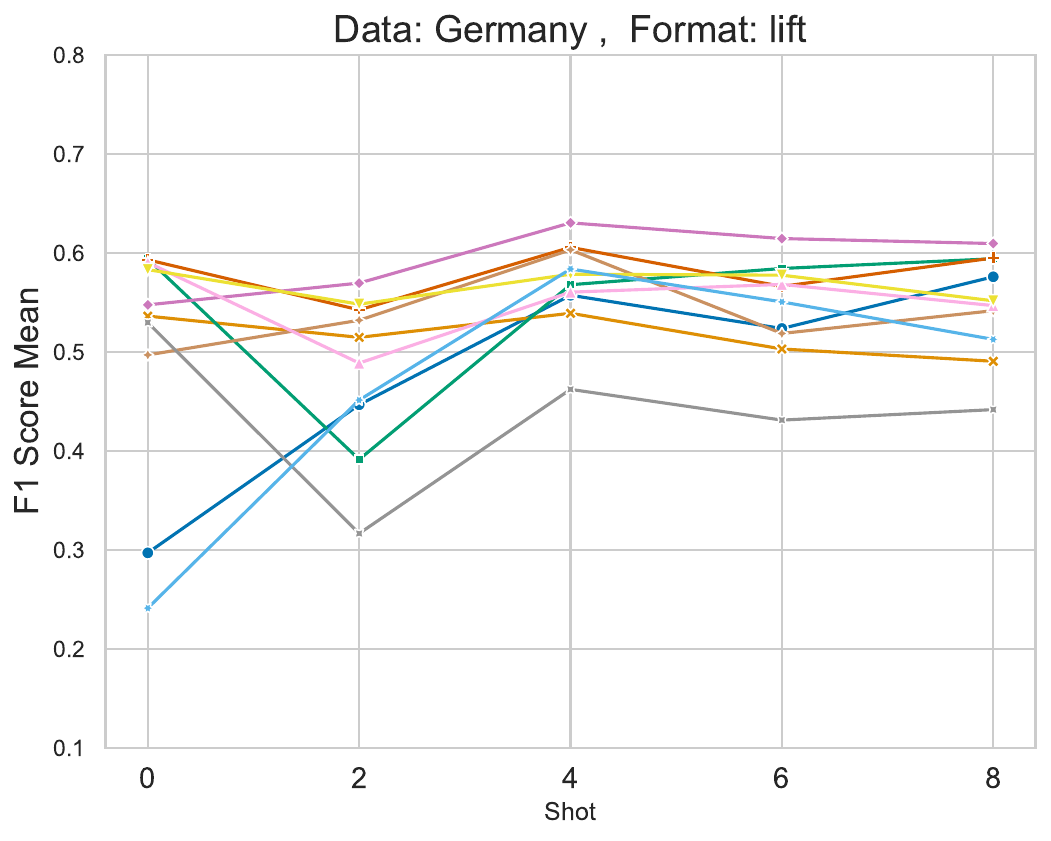}
\includegraphics[width=4.0cm]{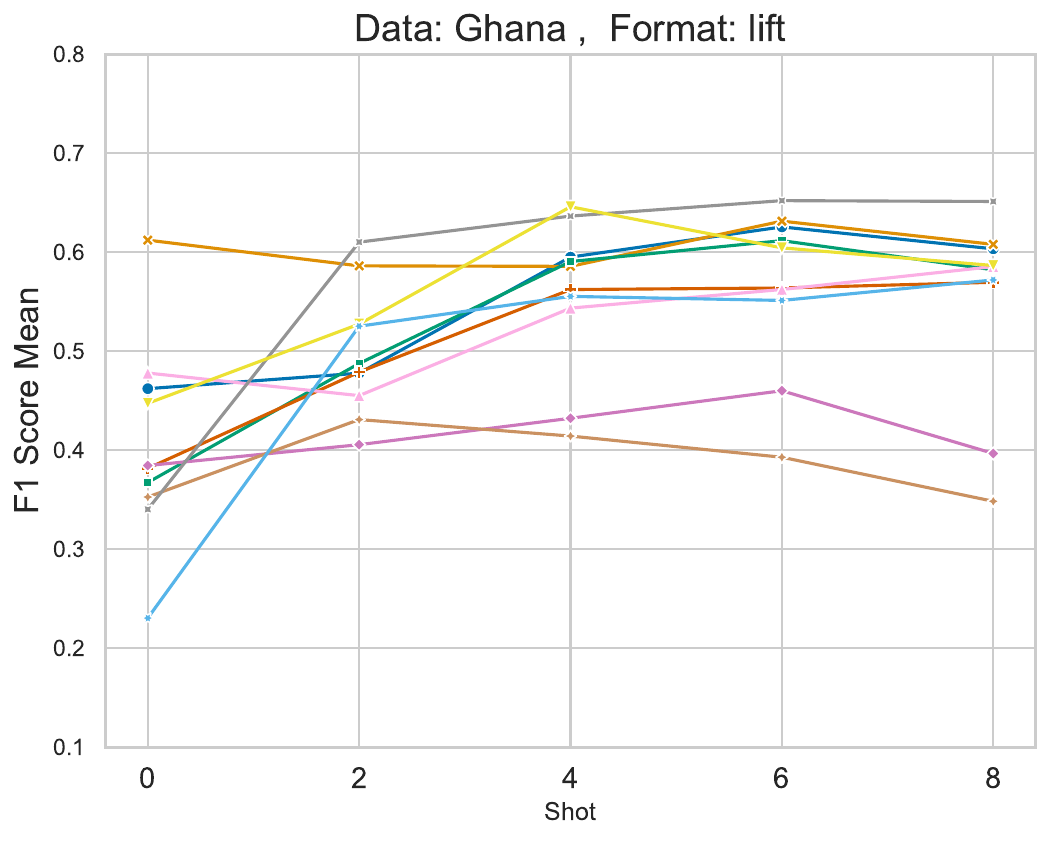}

\includegraphics[width=4.0cm]{image-pdf/f1_score_vs_shot_U.S._text.pdf}
\includegraphics[width=4.0cm]{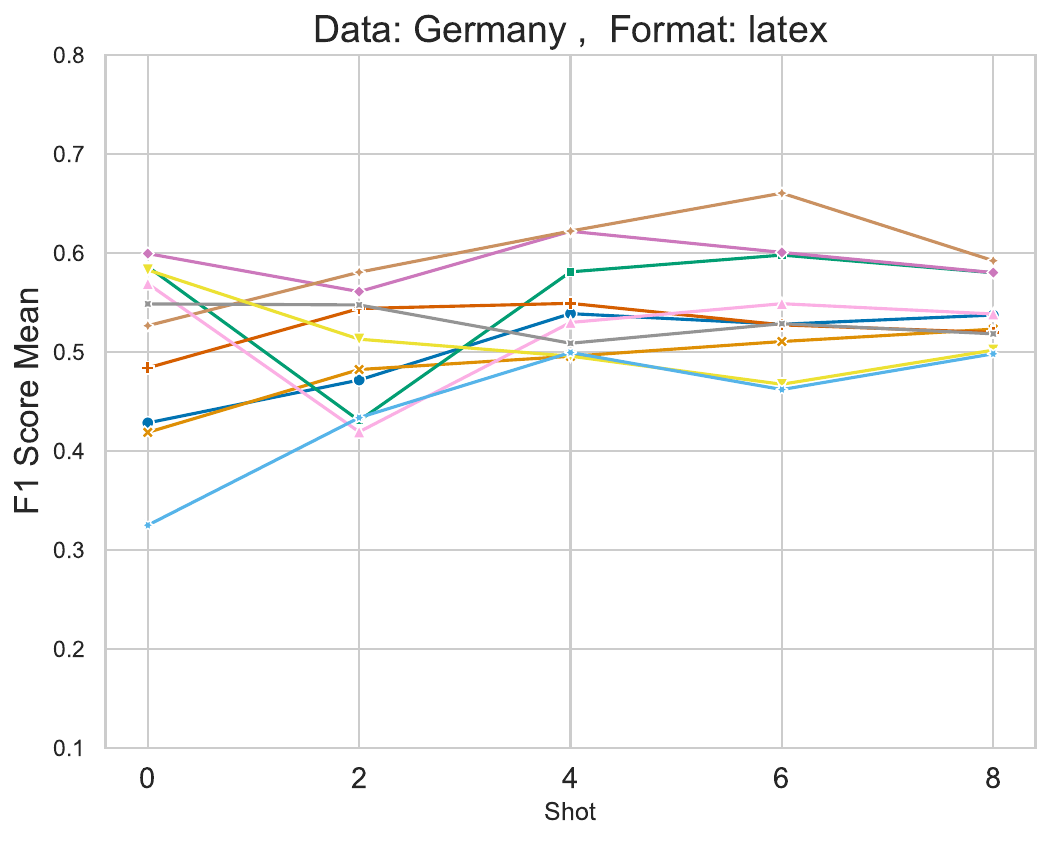}
\includegraphics[width=4.0cm]{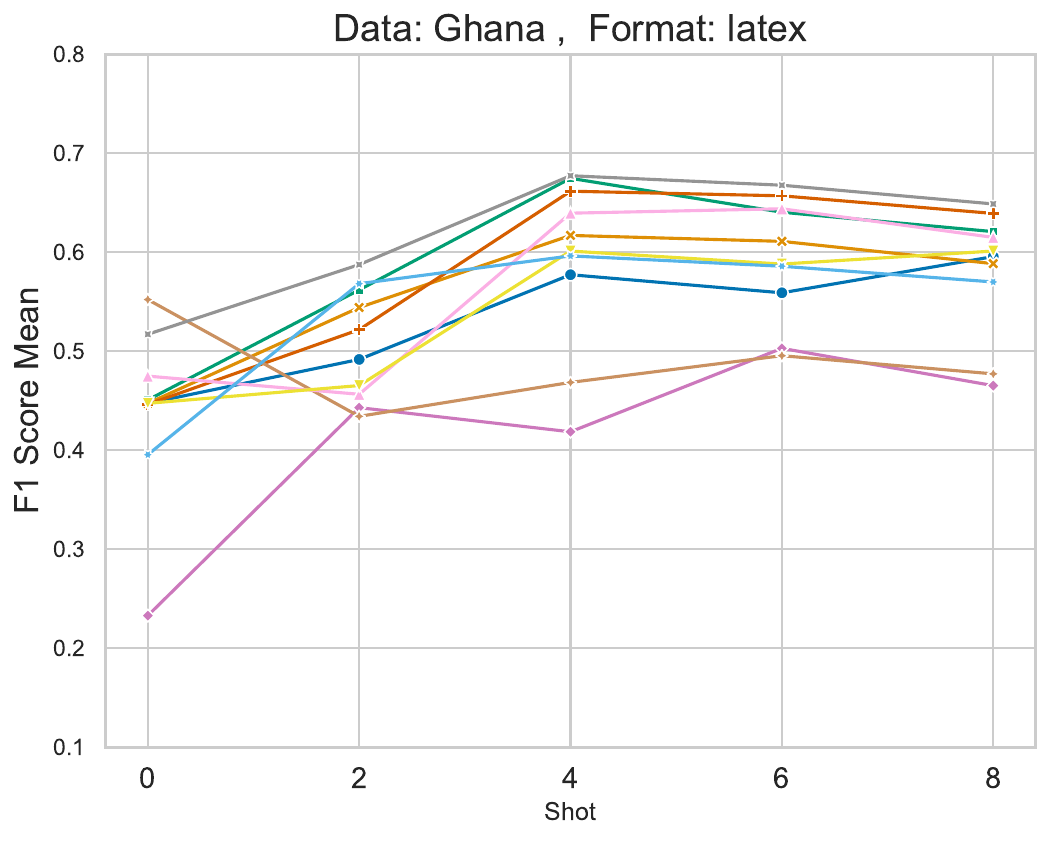}

\includegraphics[width=4.0cm]{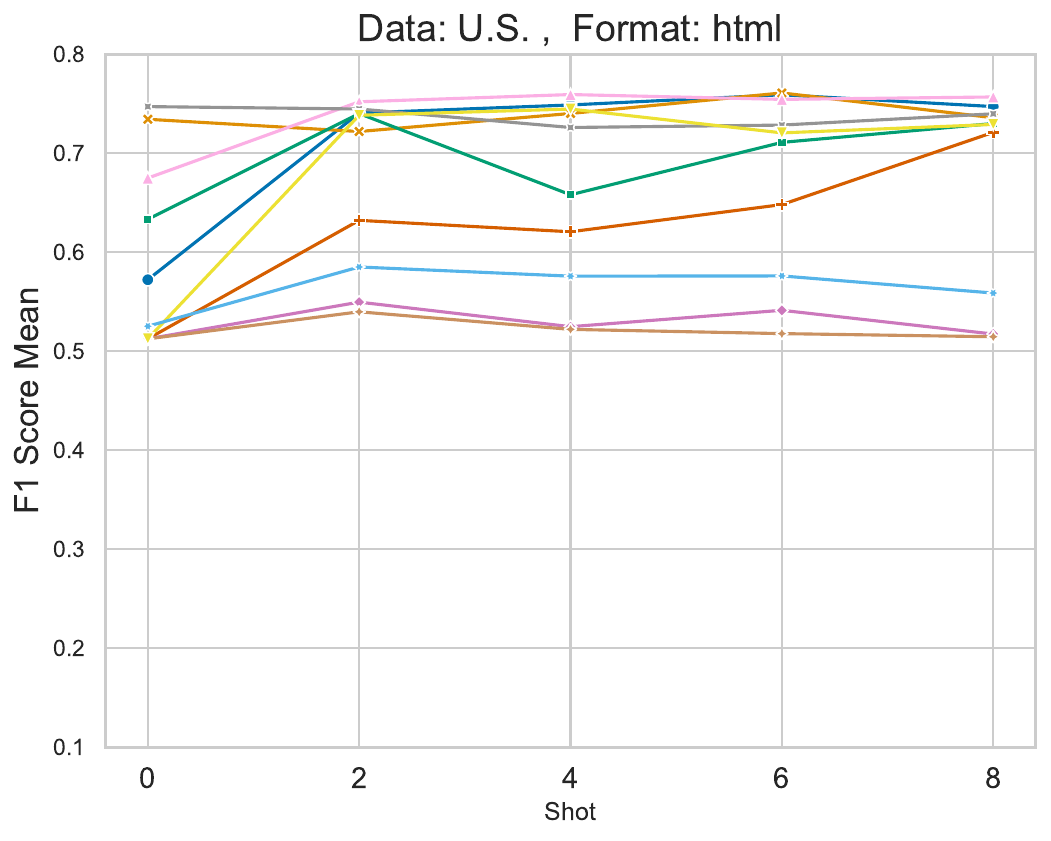}
\includegraphics[width=4.0cm]{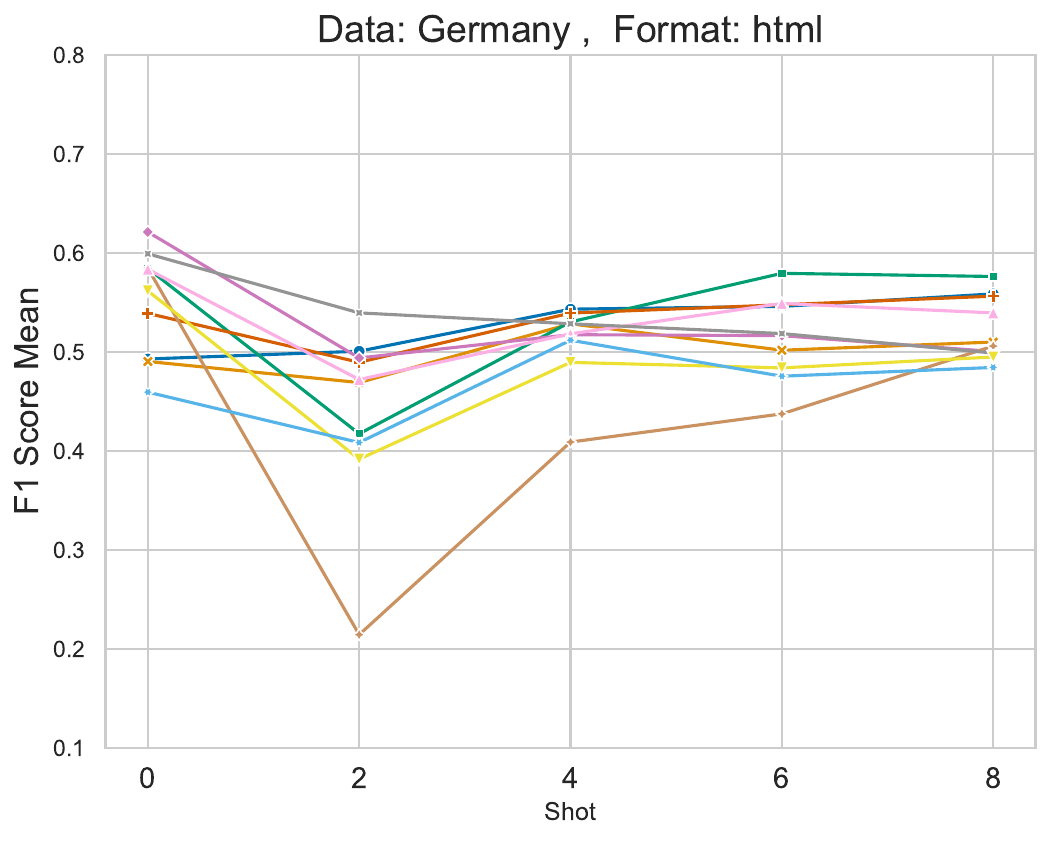}
\includegraphics[width=4.0cm]{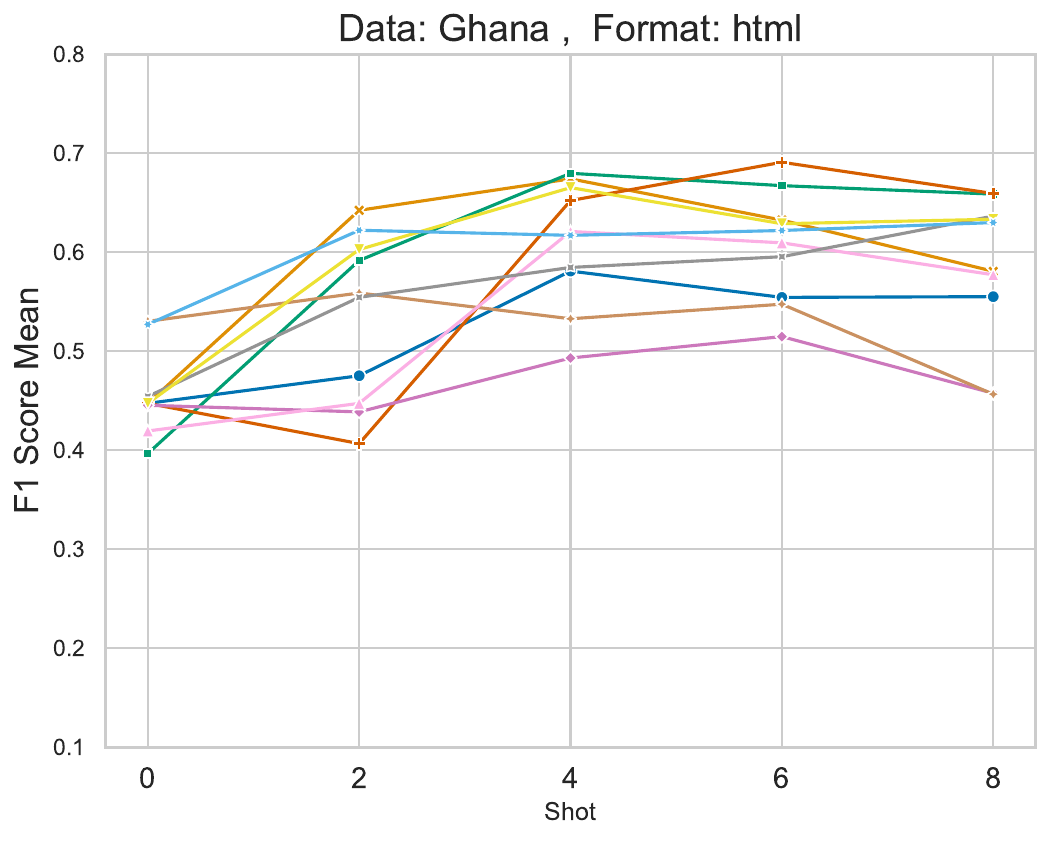}
\includegraphics[width=12cm]{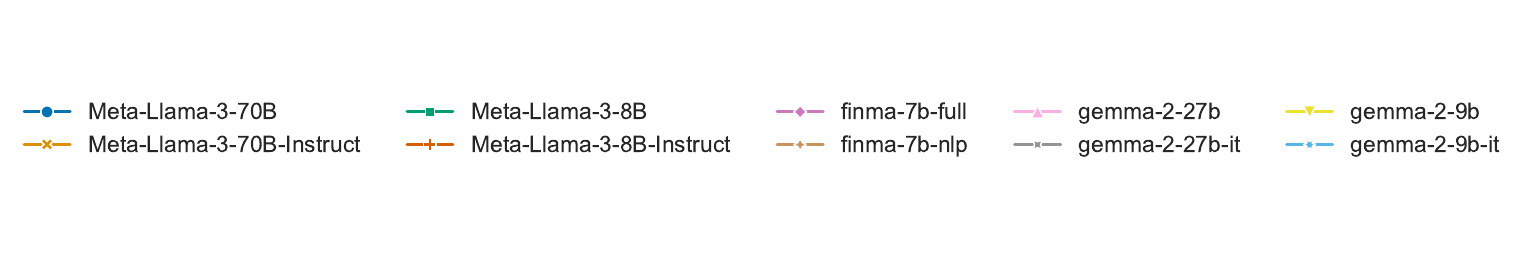}
\caption{\textbf{Average F1 Score for Few-Shot Learning Across Different Serialization Methods}
This figure presents the average F1 scores across various serialization methods for each dataset. We observe that the same models exhibit similar performance trends within each dataset, regardless of format. While the text format of the Ghana dataset may not share characteristics with the text format of the Germany dataset, Ghana’s text and JSON formats display notable similarities.  }
\label{shot-model-data}
\end{figure*}

In the In-Context Learning (ICL) experiment shown in Figure \ref{shot-model-data}, we selected balanced few-shot examples from the training set, ensuring that each set of $n$ examples was predetermined and included a balanced representation of the gender feature. Our findings indicate that ICL yields the most significant improvement when increasing from zero to two examples; however, subsequent increments in the number of examples does not result in similar returns. This observation aligns with existing research, which suggests that while ICL can be effective with a limited number of examples, its performance gains tend to plateau as more examples are added~\cite{agarwal2025many}. \\Looking at Figure \ref{shot-model-data}, we observe that decisions are more dependent on datasets than models. Particularly, finance-based models tend to show low performance in U.S. and Ghana data while \gemmanineit shows lower performance in German data.  Looking at the average across the formats \gemmatsit performs best for the U.S., \llamathree performs well for Germany.

\section{Token Attribution explainability experiments}
\label{token-attrbution-experiments}

In understanding the decision processes made by LLMs we used \emph{captum}~\cite{kokhlikyan2020captum}, an open-source model explainability
library that provides a variety of generic interpretability methods. Our main question of interest in this work was to understand the interesting features that are used by LLMs in decision-making. In addition, we seek to understand the different decision-making characteristics observed between each LLM.

In this work, the main questions we have are; if LLMs are looking at interesting attributes to make decisions and what different decision-making characteristics are observed between each LLM. 

We calculated token attribution for examples by replacing them with every possible item in the test set and assuming specific generation output. The results reported show representative values for the whole test set since we built our baseline tokens to be representative of the whole test set. Detailed visualization of the attribution is shown in Figures below.

The models explored in this study are medium-sized open-source models, chosen to balance computational efficiency and feasibility. The inclusion of larger models was limited due to computational overhead, while architectural complexities in Captum prevented the integration of financial models. 

\begin{figure*}[t]
\centering
\small{
\begin{verbatim}
    \{'sex': 1, 'amnt req': 1500, 'ration': 1, 'maturity': 30.0, 'assets val': 2000, 'dec profit': 300.0,
 'xperience': 1.0, 'educatn': 1, 'age': 53, 'collateral': 1500, 'locatn': 0, 'guarantor': 0, 'relatnshp': 1,
 'purpose': 1, 'sector': 4, 'savings': 0\}
\end{verbatim}
}
\foreach \pagenum in {1, 2, 3, 4, 5, 6, 7, 8} {
    \includegraphics[page=\pagenum,width=\linewidth]{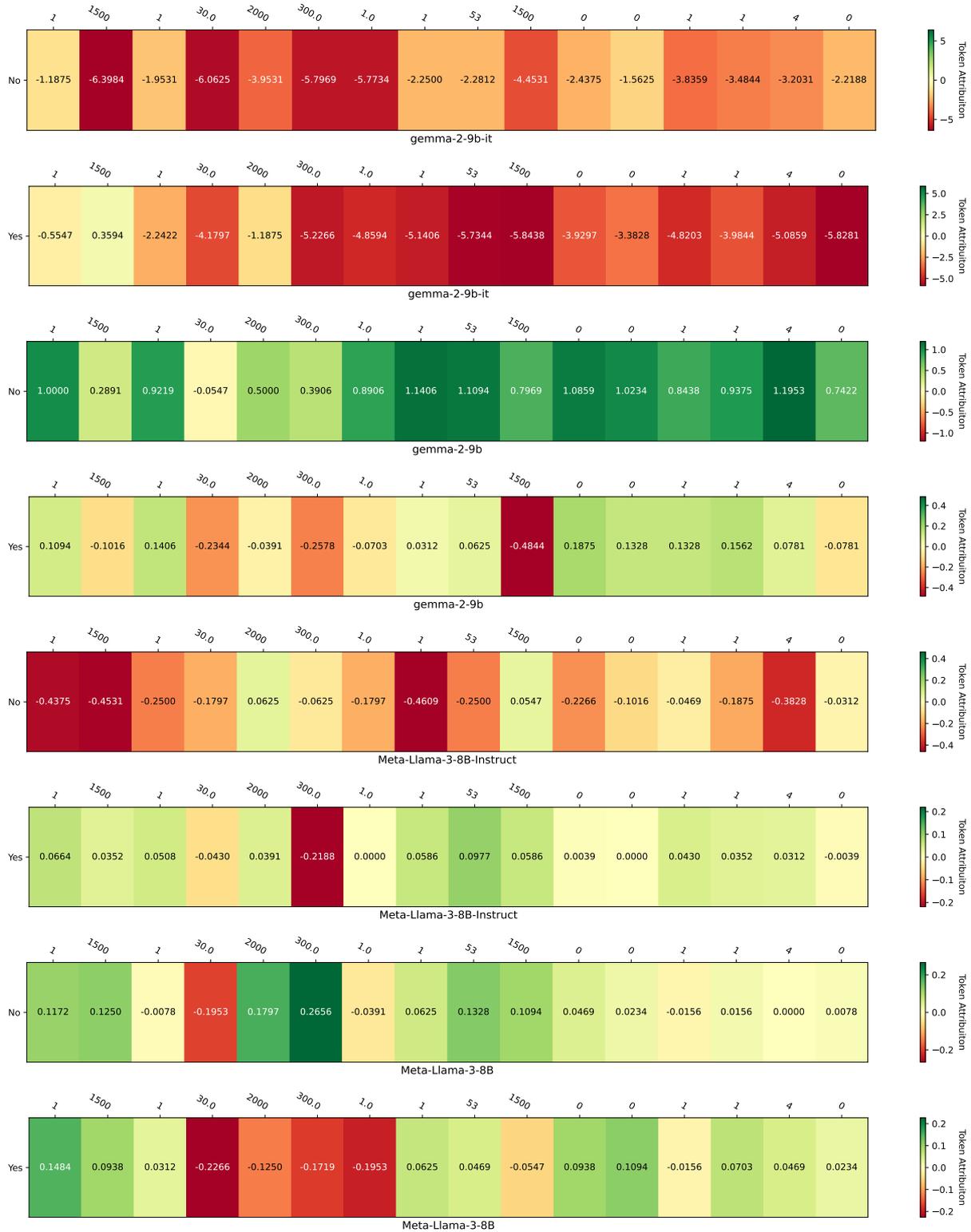}
}
\caption{Attribution scores of Ghana data for example 1. Positive attribution scores are indicated in green, while negative scores are shown in red. We can see \gemmanineit models have more negative and neutral attribution scores completely different from their original model \gemmanine .}
\label{ghana-example-1}
\end{figure*}

\begin{figure*}[t]
\centering
\small{
\begin{verbatim}
    \{'sex': 0, 'amnt req': 9000, 'ration': 0, 'maturity': 30.0, 'assets val': 10000, 'dec profit': 900.0, 
    'xperience': 3.0, 'educatn': 3,'age': 35, 'collateral': 9000, 'locatn': 1, 'guarantor': 0,
 'relatnshp': 0, 'purpose': 1, 'sector': 4, 'savings': 1\}
\end{verbatim}
}
\foreach \pagenum in {1, 2, 3, 4, 5, 6, 7, 8} {
    \includegraphics[page=\pagenum,width=\linewidth]{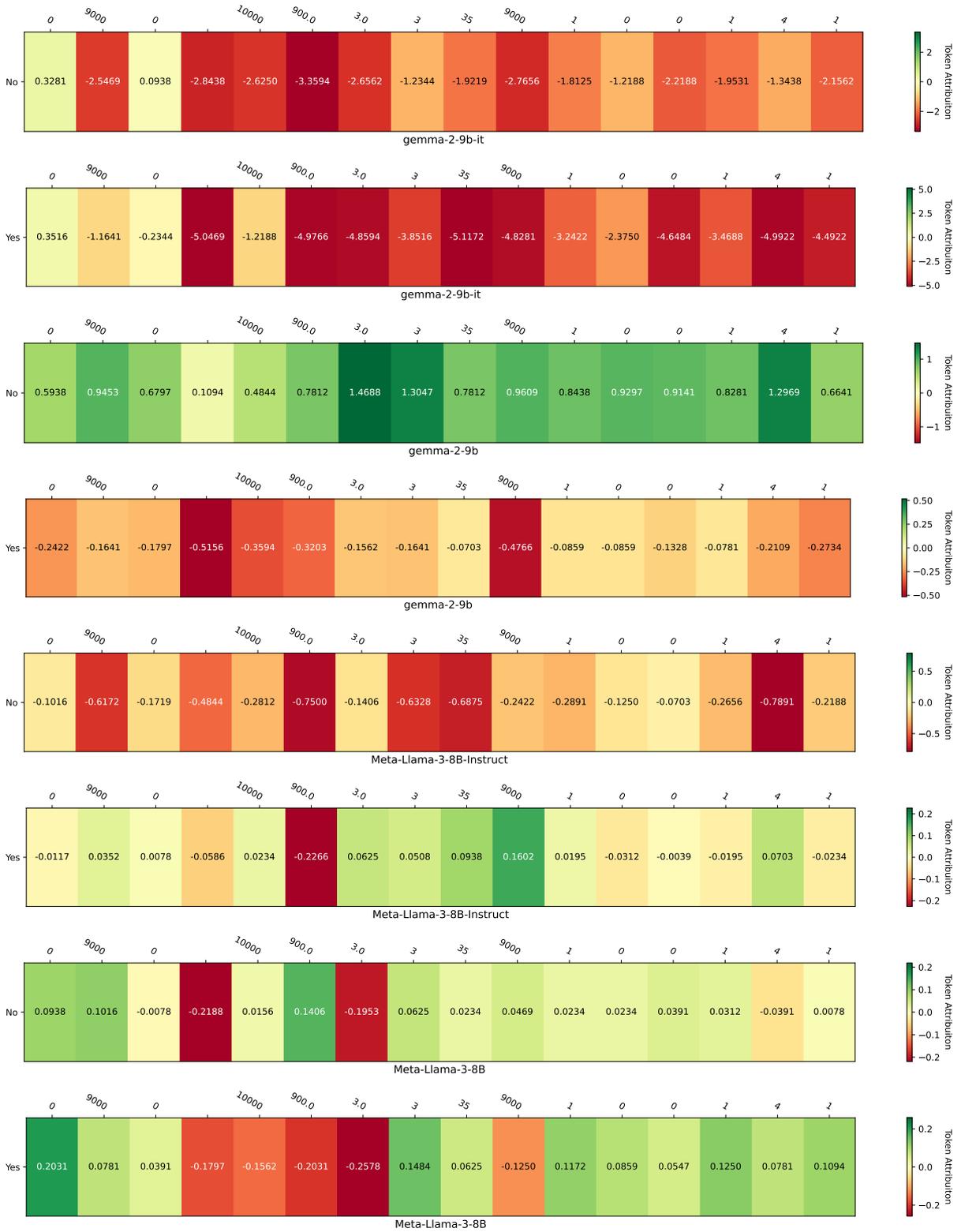}
}
\caption{Attribution scores of Ghana data for example 2. Positive attribution scores are indicated in green, while negative scores are shown in red. \gemmanineit models show more negative and neutral token attribution.}
\label{ghana-example-2}
\end{figure*}

For the Ghana dataset, as shown in Figure \ref{ghana-example-1} and Figure \ref{ghana-example-2}, we observed that \gemmanineit models primarily exhibit negative or neutral attributions from surrounding features for both positive and negative predictions. This behavior results in a slight performance gain, as presented in Table \ref{fig:zero-shot-format-serialization}. Additionally, we found no consistent feature that LLMs consistently focus on, making the decision-making process highly model-dependent.

\begin{figure*}[t]
\centering

\small{\begin{verbatim}
    \{'Gender': 'Male','Loan_ID': 'LP002101', 'Married': 'Yes','Dependents': '0', 'Education': 'Graduate',
    'Self_Employed': None, 'ApplicantIncome': 63337,'CoapplicantIncome': 0.0, 'LoanAmount': 490.0,
    'Loan_Amount_Term': 180.0, 'Credit_History': 1.0,'Property_Area': 'Urban'\}
\end{verbatim}
}
\foreach \pagenum in {1, 2, 3, 4, 5, 6, 7, 8} {
    \includegraphics[page=\pagenum,width=0.75\linewidth]{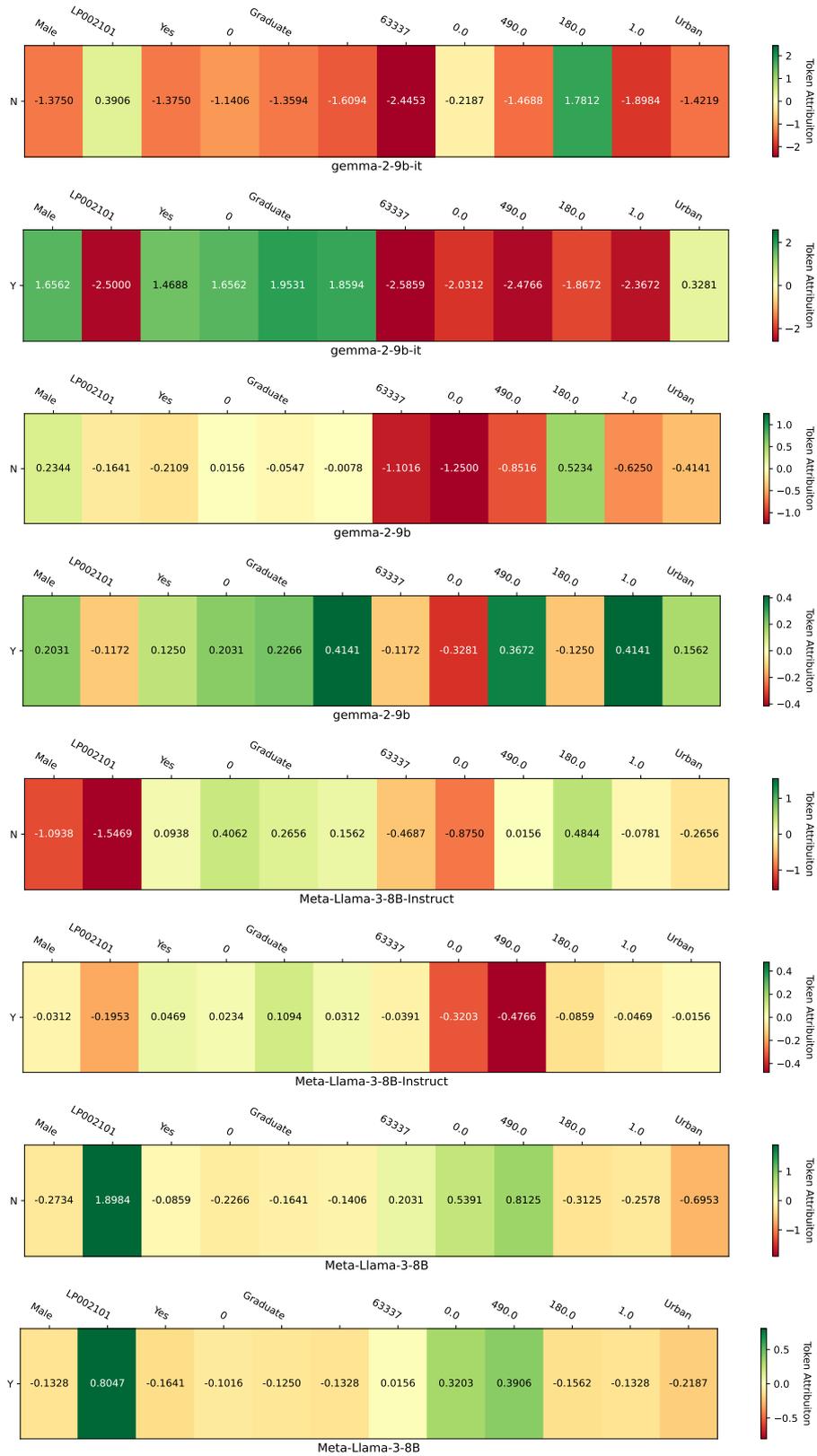}
}
\caption{This figure displays the attribution scores for Example 1 of the US dataset. Positive attribution scores are indicated in green, while negative scores are shown in red.  We can see the ``Loan\_ID'' feature significantly influences the model’s output.}
\label{attr-us-example-1}
\end{figure*}

\begin{figure*}[t]
\centering

\small{\begin{verbatim}
    \{'Gender': 'Female','Loan_ID': 'LP002978', 'Married': 'No', 'Dependents': '0','Education': 'Graduate',
'Self_Employed': 'No', 'ApplicantIncome': 2900, 'CoapplicantIncome': 0.0,'LoanAmount': 71.0, 
'Loan_Amount_Term': 360.0, 'Credit_History': 1.0, 'Property_Area': 'Rural'\}
\end{verbatim}
}
\foreach \pagenum in {1, 2, 3, 4, 5, 6, 7, 8} {
\includegraphics[page=\pagenum,width=0.75\linewidth]{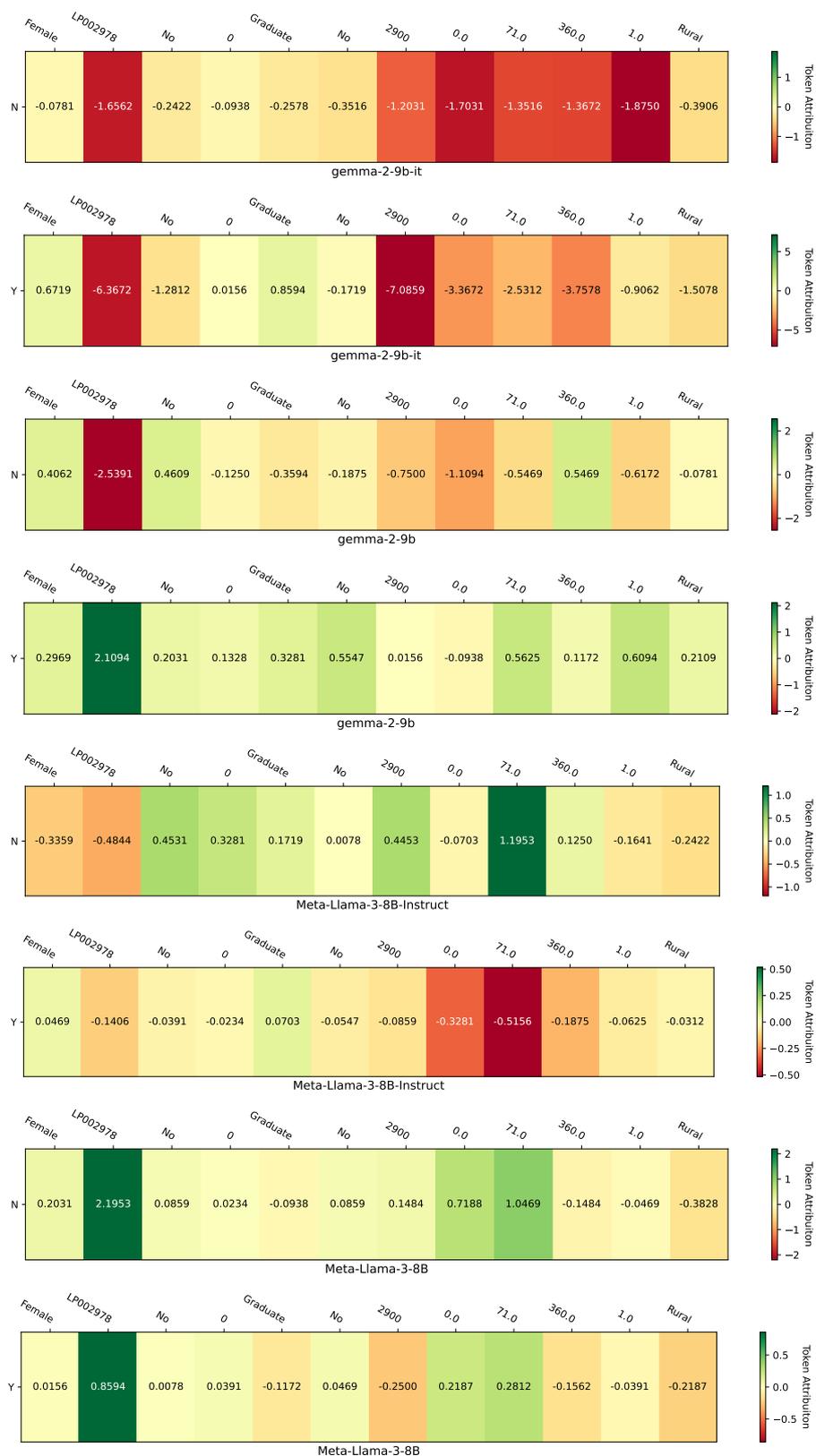}
}
\caption{This figure displays the attribution scores for Example 2 of the US dataset. Positive attribution scores are indicated in green, while negative scores are shown in red.  We can see the ``Loan\_ID'' feature significantly influences the model’s output.}
\label{attr-us-example-2}
\end{figure*}

For the US data, as shown in Figure \ref{attr-us-example-2} and Figure \ref{attr-us-example-1}, we observed that most decisions are influenced by the Loan\_ID column, which contradicts the patterns observed by manual decision-makers. Unlike other datasets, the US data exhibits more consistent feature selection by LLMs, indicating a stronger alignment in the features they prioritize.


\label{icl-result}




\begin{figure*}[t]
\centering

\small{
\begin{verbatim}
    \{'gender': 'male','checking_status': "'no checking'", 'duration': 54, 'credit_history':
    "'no credits/all paid'", 'purpose': "'used car'", 'credit_amount': 9436, 'savings_status':
    "'no known savings'", 'employment': "'1<=X<4'",'installment_commitment': 2, 'other_parties': 'none',
    'residence_since': 2, 'property_magnitude': "'life insurance'",'age': 39, 'other_payment_plans': 'none',
    'housing': 'own', 'existing_credits': 1,'job': "'unskilled resident'", 'num_dependents': 2,
    'own_telephone': 'none', 'foreign_worker': 'yes'\}
\end{verbatim}
}
\foreach \pagenum in {1, 2, 3, 4, 5, 6, 7, 8} {
    \includegraphics[page=\pagenum,width=\linewidth]{attr/germany_male.pdf}
}
\caption{This figure displays the attribution scores for Example 1 of the Germany dataset. Positive attribution scores are indicated in green, while negative scores are shown in red. \gemmanineit models show high negative attribution from most features and we don't see a focus on specific features throughout the models.}
\end{figure*}

\begin{figure*}[t]
\centering

\small{
\begin{verbatim}
    \{'gender': 'female', 'checking_status': '$<0$', 'duration': $18$, 'credit_history': 'existing paid',
    'purpose': 'radio/tv', 'credit_amount': $3190$, 'savings_status': '$<100$', 'employment': '$1 \leq X < 4$',
    'installment_commitment': $2$, 'other_parties': 'none', 'residence_since': $2$, 
    'property_magnitude': 'real estate', 'age': $24$, 'other_payment_plans': 'none', 'housing': 'own', 
    'existing_credits': $1$, 'job': 'skilled', 
    'num_dependents': $1$, 'own_telephone': 'none', 
    'foreign_worker': 'yes'\}
\end{verbatim}
}
\foreach \pagenum in {1, 2, 3, 4, 5, 6, 7, 8} {
    \includegraphics[page=\pagenum,width=\linewidth]{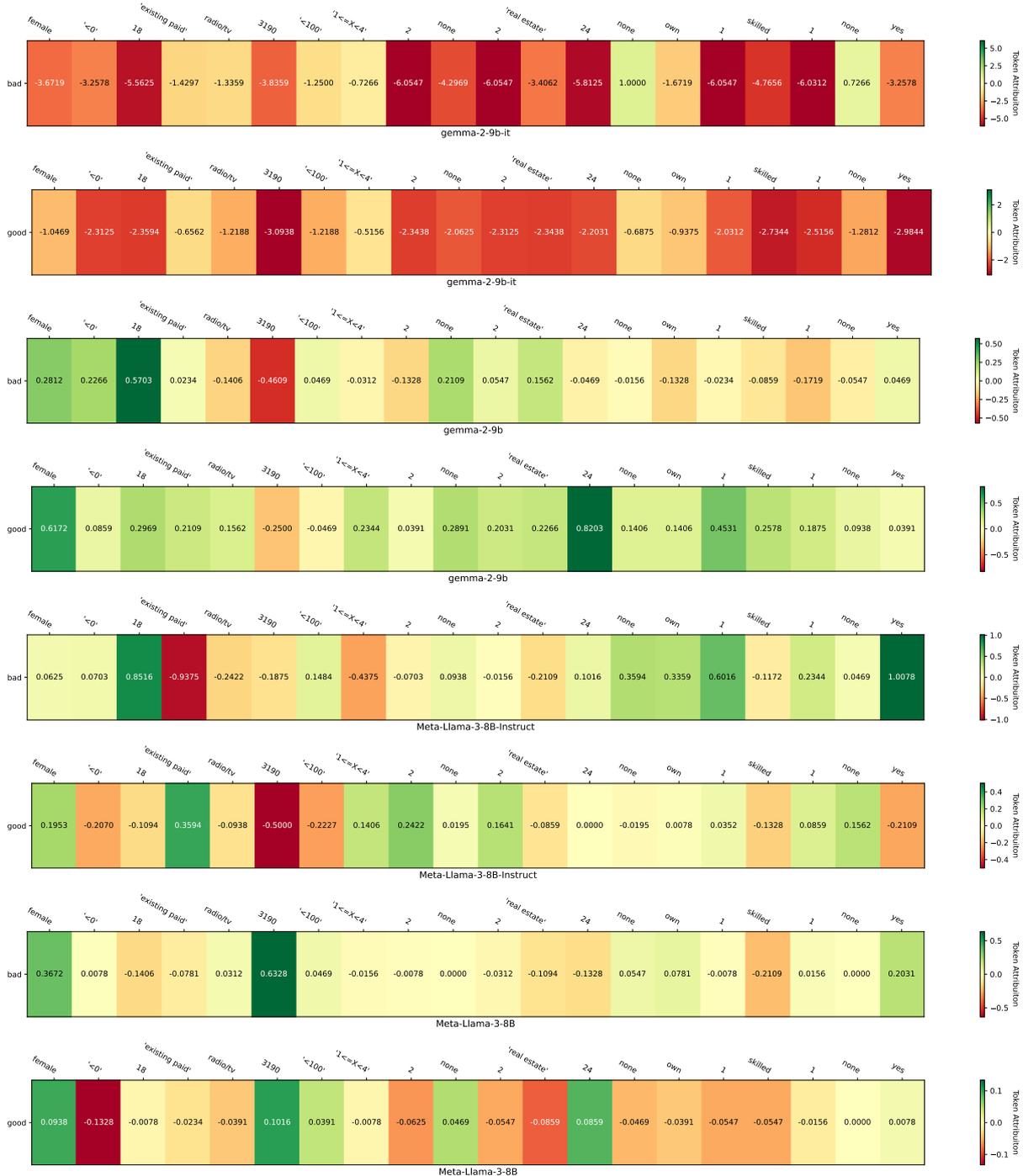}
}
\caption{This figure displays the attribution scores for Example 2 of the Germany dataset. Positive attribution scores are indicated in green, while negative scores are shown in red. \gemmanineit models show high negative attribution from most features, and we don't see a focus on specific features throughout the models.}
\end{figure*}


\end{document}